\documentclass{article}

\usepackage[final]{neurips}

\usepackage{enumitem}
\usepackage{marvosym} 
\usepackage[utf8]{inputenc} 
\usepackage[T1]{fontenc}    
\usepackage{hyperref}       
\usepackage{url}            
\usepackage{booktabs}       
\usepackage{amsfonts}       
\usepackage{microtype}      
\usepackage{xcolor}         
\usepackage{epigraph}
\usepackage[export]{adjustbox}
\usepackage{multirow}
\usepackage{listings}
\usepackage{colortbl}
\usepackage{color}
\usepackage{pifont}
\usepackage{fontawesome}
\usepackage{wrapfig}
\usepackage{subcaption}
\usepackage{caption}
\usepackage{float}
\usepackage{enumitem}
\usepackage{tabularx}
\usepackage{makecell}
\usepackage{minitoc} 

\noptcrule

\usepackage{array}

\usepackage{caption}
\usepackage{amsmath,amsfonts,bm}

\def\eqref#1{equation~\ref{#1}}

\def\1{\bm{1}}

\DeclareMathAlphabet{\mathsfit}{\encodingdefault}{\sfdefault}{m}{sl}
\SetMathAlphabet{\mathsfit}{bold}{\encodingdefault}{\sfdefault}{bx}{n}

\usepackage{xspace}
\newcommand{\ie}{\emph{i.e.,}\xspace}
\newcommand{\eg}{\emph{e.g.,}\xspace}

\newcommand{\model}[1]{\textsc{Rho-1}} 
\newcommand{\method}[1]{{SLM}} 
\newcommand{\methodfull}[1]{{Selective Language Modeling}} 

\newcommand{\mamix}[1]{\textsc{MaMix}}
\newcommand{\genmix}[1]{\textsc{GenMix}}

\makeatletter  
\renewcommand\@fnsymbol[1]{%
  \ensuremath{%
    \ifcase#1\or 
    \star\or     
    \diamond\or   
    \dagger\or  
    \mathsection\or 
    \mathparagraph\or 
    \|\or        
    **\or        
    \dagger\dagger\or 
    \ddagger\ddagger 
    \else \@ctrerr \fi%
  }%
}  
\makeatother

\definecolor{darkblue}{rgb}{0, 0, 0.5}
\hypersetup{colorlinks=true, citecolor=darkblue, linkcolor=darkblue, urlcolor=darkblue}
\definecolor{lightgray}{rgb}{0.9, 0.9, 0.9}

\definecolor{darkgreen}{RGB}{50,100,0}
\definecolor{darkred}{RGB}{200, 0, 0}
\definecolor{lightred}{RGB}{250, 200, 200}
\definecolor{lightblue}{RGB}{210, 220, 250}
\definecolor{doderblue}{RGB}{30,144,255}
\definecolor{select}{RGB}{222, 235, 247}
\definecolor{unselect}{RGB}{247, 207, 206}

\usepackage{cleveref}
\makeatletter
\AtBeginDocument{%
  \renewcommand{\sectionautorefname}{\S\@gobble}

  \renewcommand{\subsectionautorefname}{\S\@gobble}  
}
\makeatother

\title{Step-Video-T2V Technical Report: The Practice, Challenges, and Future of Video Foundation Model}

\author{Step-Video Team
\\
StepFun
}

\begin{document}

\doparttoc
\faketableofcontents

\maketitle

\begin{abstract}
\label{sec:abstract}
\vspace{-0.2cm}

We present \textbf{Step-Video-T2V}, a state-of-the-art text-to-video pre-trained model with 30B parameters and the ability to generate videos up to 204 frames in length. 
A deep compression Variational Autoencoder, \textbf{Video-VAE}, is designed for video generation tasks, achieving 16x16 spatial and 8x temporal compression ratios, while maintaining exceptional video reconstruction quality.
User prompts are encoded using two bilingual text encoders to handle both English and Chinese. A DiT with 3D full attention is trained using Flow Matching and is employed to denoise input noise into latent frames. A video-based DPO approach, \textbf{Video-DPO}, is applied to reduce artifacts and improve the visual quality of the generated videos.
We also detail our training strategies and share key observations and insights. Step-Video-T2V's performance is evaluated on a novel video generation benchmark, \textbf{Step-Video-T2V-Eval}, demonstrating its state-of-the-art text-to-video quality when compared with both open-source and commercial engines. Additionally, we discuss the limitations of current diffusion-based model paradigm and outline future directions for video foundation models. We make both Step-Video-T2V and Step-Video-T2V-Eval available at \hyperlink{https://github.com/stepfun-ai/Step-Video-T2V}{https://github.com/stepfun-ai/Step-Video-T2V}. The online version can be accessed from \hyperlink{https://yuewen.cn/videos}{https://yuewen.cn/videos} as well. Our goal is to accelerate the innovation of video foundation models and empower video content creators. 

\end{abstract}

\section{Preface}

A video foundation model is a model pre-trained on large video datasets that can generate videos in response to text, visual, or multimodal inputs from users. It can be applied to a wide range of downstream video-related tasks, such as text/image/video-to-video generation, video understanding and editing, as well as video-based conversion, question answering, and task completion.

Based on our understanding, we define two levels towards building video foundation models. 
\textbf{Level-1: translational video foundation model}. A model at this level functions as a cross-modal translation system, capable of generating videos from text, visual, or multimodal context.
\textbf{Level-2: predictable video foundation model}. A model at this level acts as a prediction system, similar to large language models (LLMs), that can forecast future events based on text, visual, or multimodal context and handle more advanced tasks, such as reasoning with multimodal data or simulating real-world scenarios.

Current diffusion-based text-to-video models, such as Sora \cite{openaisora}, Veo \cite{veo}, Kling \cite{kling}, Hailuo \cite{hailuo}, and Step-Video (as described in this report), belong to Level-1. These models can generate high-quality videos from text prompts, lowering the barrier for creators to produce video content. However, they often fail to generate videos that require complex action sequences (such as a gymnastic performance) or adherence to the laws of physics (such as a basketball bouncing on the floor), let alone performing causal or logical tasks like LLMs. Such limitations arise because these models learn only the mappings between text prompts and corresponding videos, without explicitly modeling the underlying causal relationships within videos. Autoregression-based text-to-video models introduce the causal modeling mechanism by predicting the next video token, frame, or clip. However, these models still cannot achieve performance comparable to diffusion-based models on text-to-video generation.

This report will detail the practice of building Step-Video-T2V as a state-of-the-art video foundation model at Level-1. By analyzing the challenges identified through experiments, we will also highlight key problems that need to be addressed in order to develop video foundation models at Level-2.
\section{Introduction}

Large language models (LLMs), as part of Artificial General Intelligence (AGI), has made impressive progress in recent years. These models are capable of understanding human instructions and generating coherent, fluent responses in natural language. However, language is a symbolic abstraction of thought, using words and concepts to represent the world. This abstraction often falls short in capturing the complexity and richness of reality, particularly when it comes to dynamic processes like object motion or the intricate spatial and temporal relationships between entities.
As a result, video generation has emerged as an important frontier in the pursuit of AGI, offering a pathway toward bridging these cognitive gaps. Moreover, video content is now the dominant form of communication and entertainment online. Developing video generation systems capable of producing high-quality content can significantly reduce barriers for creators and democratize video production. This empowers everyone, from amateurs to professionals, to effortlessly create compelling videos.

In this technical report, we present Step-Video-T2V, a state-of-the-art video foundation model with 30B parameters, capable of generating high-quality videos from text, featuring strong motion dynamics, high aesthetics, and consistent content. Like most commercial video generation engines, Step-Video-T2V is a diffusion Transformer (DiT)-based model trained using Flow Matching. A specially designed deep compression Variational Auto-encoder (VAE) achieves 16x16 spatial and 8x temporal compression ratios, significantly reducing the computational complexity of large-scale video generation training. Two bilingual text encoders enable Step-Video-T2V to directly understand Chinese or English prompts. A cascaded training pipeline, including text-to-image pre-training, text-to-video pre-training, supervised fine-tuning (SFT), and direct preference optimization (DPO), is introduced to accelerate model convergence and fully leverage video datasets of varying quality. A new benchmark dataset called Step-Video-T2V-Eval is created for text-to-video generation, which includes 128 diverse prompts across 11 categories, alongside video generation results from several top text-to-video open-source and commercial engines for comparison.

Insights are gained throughout the entire development of Step-Video-T2V, spanning data, model, training, and inference. First, text-to-image pre-training is essential for the video generation model to acquire rich visual knowledge, including concepts, scenes, and their spatial relationships, providing a solid foundation for the subsequent text-to-video pre-training stages. Second, text-to-video pre-training at low resolution is critical for the model to learn motion dynamics. The more stable the model is trained during this stage, using as much diverse training data as possible, the easier it becomes to refine and scale the model to higher resolutions and more complex video generation tasks. Third, using high-quality videos with accurate captions and desired styles in SFT is crucial to the stability of the model and the style of the generated videos. Fourth, video-based DPO can further enhance the visual quality by reducing artifacts, ensuring smoother and more realistic video outputs.

Challenges remain in state-of-the-art video foundation models. For example, current video captioning models still face hallucination issues, leading to unstable training and poor instruction-following performance. Composing multiple concepts with low occurrence in the training data (e.g., an elephant and a penguin) within a single generated video is still a difficult task. Additionally, training and generating long-duration, high-resolution videos still face significant computational cost hurdles. Furthermore, even a DiT-based model like Step-Video-T2V with 30B parameters struggles to generalize well when generating videos involving complex action sequences or requiring adherence to the laws of physics. By open-sourcing Step-Video-T2V, we aim to provide researchers and engineers with a strong baseline, helping them better understand these challenges and accelerate innovations in the development and application of video foundation models.

The key contributions of this technical report are as follows:
\begin{itemize}[left=0cm] 
\item We present and open-source Step-Video-T2V, a state-of-the-art  text-to-video pre-trained model with 30B parameters, capable of understanding both Chinese and English prompts, generating high-quality videos (544x992 resolution) up to 204 frames in length, featuring strong motion dynamics, high aesthetics, and consistent content.
\item We introduce a deep compression Video-VAE for video foundation models, achieving 16x16 spatial and 8x temporal compression ratios, while maintaining exceptional video reconstruction quality.
\item We detail the optimizations of model hyper-parameters, operators, and parallelism in Step-Video-T2V, which ensure both the stability and efficiency of training from a system-level perspective.
\item We describe the process of pre-processing large-scale videos as training data, and explain how these videos are filtered and utilized at different stages of training.
\item We release Step-Video-T2V-Eval as a new benchmark, which includes 128 diverse prompts across 11 categories and video generation results from top open-source and commercial engines.
\item We discuss the insights and challenges encountered in the development of Step-Video-T2V, and identify key issues that must be addressed to advance video foundation models.
\end{itemize}

\section{Related Work}
Video generation technology has seen significant progress over the past year, with advancements from Sora \cite{openaisora} to Gen-3 \cite{runwaygen3}, Kling \cite{kling}, Hailuo \cite{hailuo}, Veo \cite{veo}, and others. 

Commercial video generation engines (e.g., Sora, Gen-3, Kling, and Hailuo) offer text-to-video generation capabilities, as well as extended applications like image-to-video generation or specialized video effect generation. Compared to these closed-source engines, which often involve longer and more complex video generation pipelines with extensive pre- and post-processing, Step-Video-T2V delivers comparable performance for general text prompts and even surpasses them in specific domains, such as generating videos with high motion dynamics or text content.

Open-source video generation models, such as HunyuanVideo \cite{kong2024hunyuanvideo}, CogVideoX \cite{yang2024cogvideox}, Open-Sora \cite{opensora}, and Open-Sora-Plan \cite{lin2024open}, offer greater transparency in their implementations, making them more accessible to researchers and content creators. Both HunyuanVideo and CogVideoX are based on MMDiT \cite{esser2024scalingrectifiedflowtransformers}, a variation of the full attention Transformer architecture. Open-Sora and Open-Sora-Plan are built on DiT \cite{peebles2023scalablediffusionmodelstransformers}, with the former using spatial-temporal attention and the latter employing full attention. Compared to these open-source models, the key contributions of Step-Video-T2V include being the largest open-source model to date, utilizing a high-compression VAE for videos, supporting bilingual text prompts in both English and Chinese, implementing a video-based DPO approach to further reduce artifacts and enhance visual quality, and providing comprehensive training and inference documentation, as outlined in this report.

Movie Gen Video \cite{polyak2024moviegencastmedia} is another video generation model from Meta, featuring a similar architecture and model size. Compared to Movie Gen Video, Step-Video-T2V stands out with four unique features. First, it incorporates a more powerful high compression VAE for large-scale video generation training. Second, it supports bilingual text prompt understanding in both English and Chinese. Third, it adds an additional DPO stage to the training process, reducing
artifacts and improving the visual quality of the generated videos. Fourth, it is open-source and provides state-of-the-art video generation quality comparing with both open-source and commercial engines.

Videos encompass both spatial and temporal information, leading to significantly larger data volumes compared to images. Addressing the computational challenge of modeling video data efficiently is therefore a fundamental problem. Various methods have been proposed to reduce the complexity of video modeling. These include approaches such as 3D Causal Convolution~\citep{yu2024language,yang2024cogvideox,kong2024hunyuanvideo,opensora}, wavelet transform~\citep{agarwal2025cosmos,li2024wf}, and Residual Autoencoding in images~\citep{chen2025deep}. While these methods show promise in terms of either reconstruction quality or compression ratio, achieving a balance between high quality and effective compression remains difficult. Our work addresses this challenge, providing a solution that opens new possibilities in video generation, such as extending the context length or scaling up the DiT model more aggressively.
\section{Model}

\begin{figure*}[t]
    \centering
    \includegraphics[width=\textwidth]{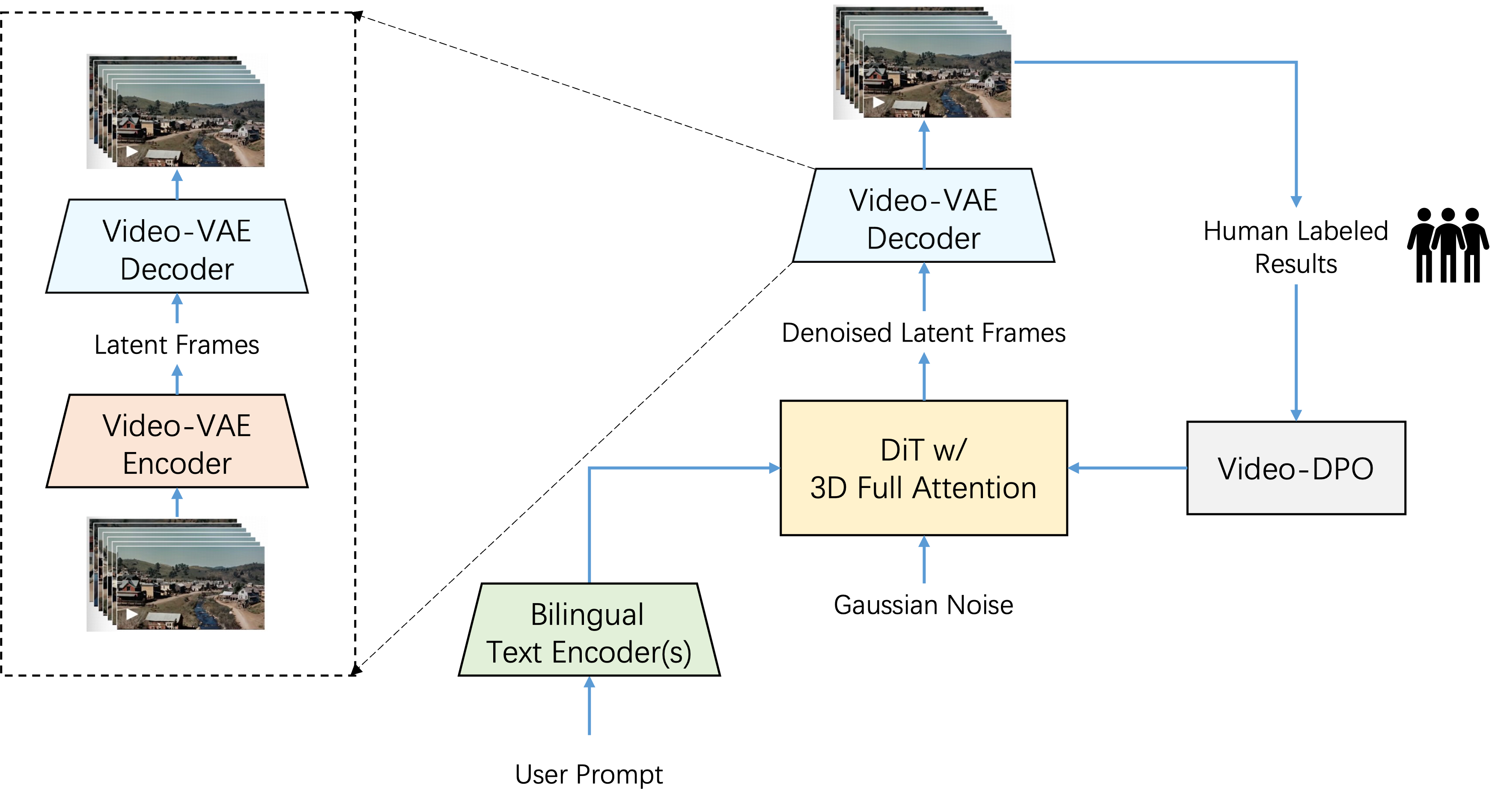}
    \caption{Architecture overview of Step-Video-T2V. Videos are represented by a high-compression Video-VAE, achieving 16x16 spatial and 8x temporal compression ratios. User prompts are encoded using two bilingual pre-trained text encoders to handle both English and Chinese. A DiT with 3D full attention is trained using Flow Matching and is employed to denoise input noise into latent frames, with text embeddings and timesteps serving as conditioning factors. To further enhance the visual quality of the generated videos, a video-based DPO approach is applied, which effectively reduces artifacts and ensures smoother, more realistic video outputs.}
    \label{fig:overview}
\end{figure*}

The overall architecture of Step-Video-T2V is given in Figure~\ref{fig:overview}. Videos are represented by a high-compression Video-VAE, achieving 16x16 spatial and 8x temporal compression ratios. User prompts are encoded using two bilingual pre-trained text encoders to handle both English and Chinese. A DiT with 3D full attention is trained using Flow Matching \citep{lipman2023flowmatchinggenerativemodeling} and is employed to denoise input noise into latent frames, with text embeddings and timesteps serving as conditioning factors. To further enhance the visual quality of the generated videos, a video-based DPO approach is applied, which effectively reduces artifacts and ensures smoother, more realistic video outputs.

Next, we will introduce the implementation details of Video-VAE, bilingual text encoders, DiT with 3D full attention, and Video-DPO, respectively.

\subsection{Video-VAE}
\subsubsection{Latent Space Compression in Video Generation}

State-of-the-art video models, such as HunyuanVideo \citep{kong2024hunyuanvideo}, CogVideoX \citep{yang2024cogvideox}, and Meta Movie Gen \citep{polyak2024moviegencastmedia}, leverage Variational Autoencoders (VAEs) with spatial-temporal downscaling factors of $4{\times}8{\times}8$ or $8{\times}8{\times}8$. These VAEs map 3-channel RGB inputs to 16-channel latent representations, achieving compression ratios as high as 1:96. To further reduce the number of tokens, these systems typically employ patchifiers that group $2{\times}2{\times}1$ latent patches into individual tokens. 

While this two-stage process of compression and tokenization is effective, it introduces architectural complexity and can potentially degrade the performance of the subsequent diffusion stages. The efficiency of text-to-video diffusion-transformer models is fundamentally dependent on their ability to operate within compressed latent spaces. Given that computational costs scale quadratically with the number of tokens due to attention operations, it is crucial to mitigate spatial-temporal redundancy through effective compression. This not only accelerates training and inference but also aligns with the diffusion process's inherent preference for condensed representations.

\subsubsection{Advancing Compression through New Architecture}
\begin{figure*}[t]
    \centering
    \includegraphics[width=0.7\textwidth, clip, trim=0.0cm 2.5cm 7.6cm 2.6cm,]{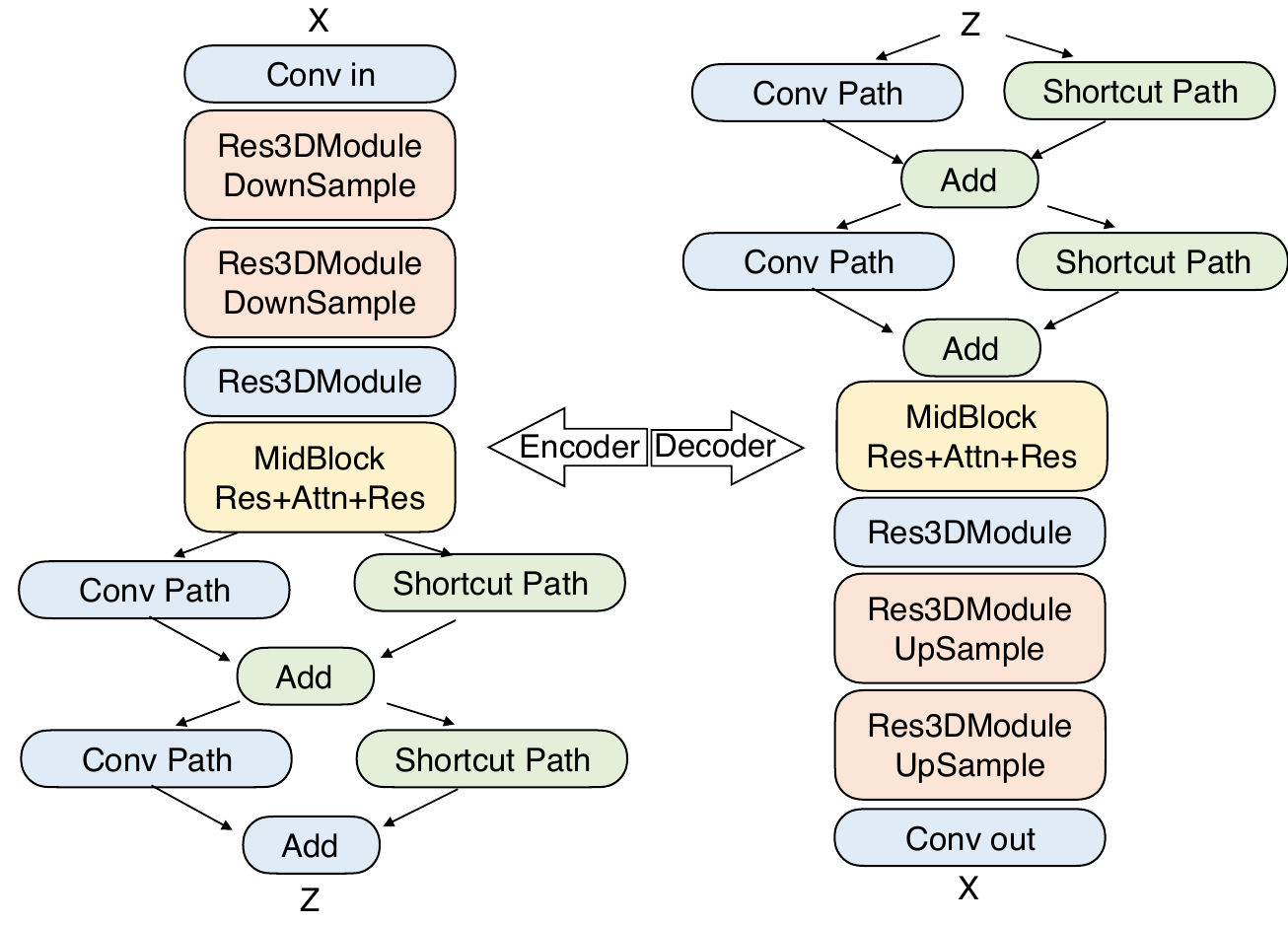}
    \caption{Architecture overview of Video-VAE.}
    \label{fig:vae}
\end{figure*}

Our Video-VAE introduces a novel dual-path architecture at the later stage of the encoder and the early stage of the decoder, featuring unified spatial-temporal compression. This design achieves $8{\times}16{\times}16$ downscaling through the synergistic use of 3D convolutions and optimized pixel unshuffling operations.
For an input video tensor $\mathbf{X} \in \mathbb{R}^{B \times C \times T \times H \times W}$, the encoder $E$ produces latent representation $\mathbf{Z} = E(\mathbf{X}) \in \mathbb{R}^{B \times C_z \times \lceil T/8 \rceil \times \lceil H/16 \rceil \times \lceil W/16 \rceil}$ through:

\paragraph{Causal 3D Convolutional Modules}
The early stage of the encoder consists of three stages, each featuring two Causal Res3DBlock and corresponding downsample layers. Following this, a MidBlock combines convolutional layers with attention mechanisms to further refine the compressed representations.
To enable joint image and video modeling, we employ temporal causal 3D convolution.
Our architecture implements temporal causality through:
\begin{equation}
    \mathcal{C}_{3D}(\mathbf{X})_t = 
    \begin{cases} 
        \text{Conv3D}([\mathbf{0},...,\mathbf{X}_{t}], \mathbf{\Theta}) & t=0 \\
        \text{Conv3D}([\mathbf{X}_{t-k},...,\mathbf{X}_t], \mathbf{\Theta}) & t>0 
    \end{cases}
\end{equation}

where $k$ is the temporal kernel size, ensuring frame $t$ only depends on previous frames.

\paragraph{Dual-Path Latent Fusion} 
\textit{The primary motivation for Dual-Path Latent Fusion is to maintain high-frequency details through convolutional processing while preserving low-frequency structure via channel averaging.} Notably, \cite{chen2025deep} identify similar mechanisms within the realm of image VAE modeling. Our approach, however, introduces a unified structure adept at handling both image and video data. This approach allows the network to use its parameters more efficiently, thereby overcoming the blurring artifacts typically associated with traditional VAEs.
\begin{enumerate}[left=0pt]
    \item \textbf{Conv Path}: Combines causal 3D convolutions with pixel unshuffling,
    \begin{equation}     
        \mathbf{H}_{\text{conv}} = \mathcal{U}^{(3)}_s\left(\mathcal{C}_{3D}(\mathbf{X})\right)
    \end{equation}

    where $\mathcal{U}^{(3)}_s: \mathbb{R}^{B \times C \times T \times H \times W} \to  \mathbb{R}^{B \times C \cdot s^3 \times \frac{T}{s_t} \times \frac{H}{s_s} \times \frac{W}{s_s}} $ with spatial stride $s_s=2$, temporal stride $s_t=2$, and $\mathcal{C}_{3D}$ denoting our causal 3D convolution.

    \item \textbf{Shortcut Path}: Preserves structural semantics through grouped channel averaging,
    \begin{equation} 
        \mathbf{H}_{\text{avg}} = \frac{1}{s^3}\sum_{k=0}^{s^3-1}\mathcal{U}^{(3)}_s(\mathbf{X})_{[...,kC_z:(k+1)C_z]}
    \end{equation}
    where $\mathcal{U}^{(3)}_s$ implements 3D pixel unshuffle with spatial-temporal blocking, $C_z$ is the latent dim of next stage.
\end{enumerate}

The output of fusion combines both paths through residual summation:
\begin{equation}  
    \mathbf{Z} = \mathbf{H}_{\text{conv}} \oplus \mathbf{H}_{\text{avg}}
\end{equation}

\subsubsection{Decoder Architecture}

The early stage of the decoder consists of two symmetric Dual Path architectures. In these architectures, the 3D pixel unshuffle operation $\mathcal{U}$ is replaced by 3D pixel shuffle operator $\mathcal{P}$, the grouped channel averaging path is replaced by a grouped channel repeating operation, which efficiently unfolds the compressed information into spatial-temporal dimensions.
In ResNet backbone, we replace all groupnorm with spatial groupnorm to avoid temporal flickering between different chunks.

\subsubsection{Training Details}

\textit{Our VAE training process is meticulously designed in multiple stages, which is the key reason for achieving our final goal of efficient and high-quality video data modeling.}

In the first stage, we train a VAE with a 4x8x8 compression ratio without employing a dual-path structure. This initial training is conducted jointly on images and videos of varying frame counts, adhering to a preset ratio. In this stage, we set a lower compress goal for model to sufficiently learn low level representations.

In the second stage, we enhance the model by incorporating two dual-path modules in both the encoder and decoder, replacing the latter part after the mid-block. During this phase, we gradually unfreeze the dual-path modules, the mid-block, and the ResNet backbone, allowing for a more refined and flexible training process.

Throughout the training, we utilize a combination of L1 reconstruction loss, Video-LPIPS, and KL-divergence constrain to guide the model. Once these losses have converged, we introduce GAN loss to further refine the model's performance. This staged approach ensures a robust and high-quality VAE capable of handling complex video data efficiently.

\subsection{Bilingual Text Encoder}\label{textencoder}
The text encoder plays a crucial role in text-to-video generation by guiding the model in the latent space. In Step-Video-T2V, we use two bilingual text encoders to process user text prompts: Hunyuan-CLIP and Step-LLM.

Hunyuan-CLIP is the bidirectional text encoder of an open-source bilingual CLIP model \cite{li2024hunyuanditpowerfulmultiresolutiondiffusion}. Due to the training mechanism of the CLIP model, Hunyuan-CLIP can produce text representations well-aligned with the visual space. However, because its maximum input length is limited to 77 tokens, Hunyuan-CLIP faces challenges when processing longer user prompts.

Step-LLM, on the other hand, is an in-house, unidirectional bilingual text encoder pre-trained using the next-token prediction task. It incorporates a redesigned Alibi-Positional Embedding \cite{press2022trainshorttestlong} to improve both efficiency and accuracy in sequence processing. Unlike Hunyuan-CLIP, Step-LLM has no input length restriction, making it particularly effective for handling lengthy and complex text sequences.

By combining these two text encoders, Step-Video-T2V is able to handle user prompts of varying lengths, generating robust text representations that effectively guide the model in the latent space.

\subsection{DiT w/ 3D Full Attention}

\begin{figure}[h]
    \centering
    \includegraphics[width=\textwidth]{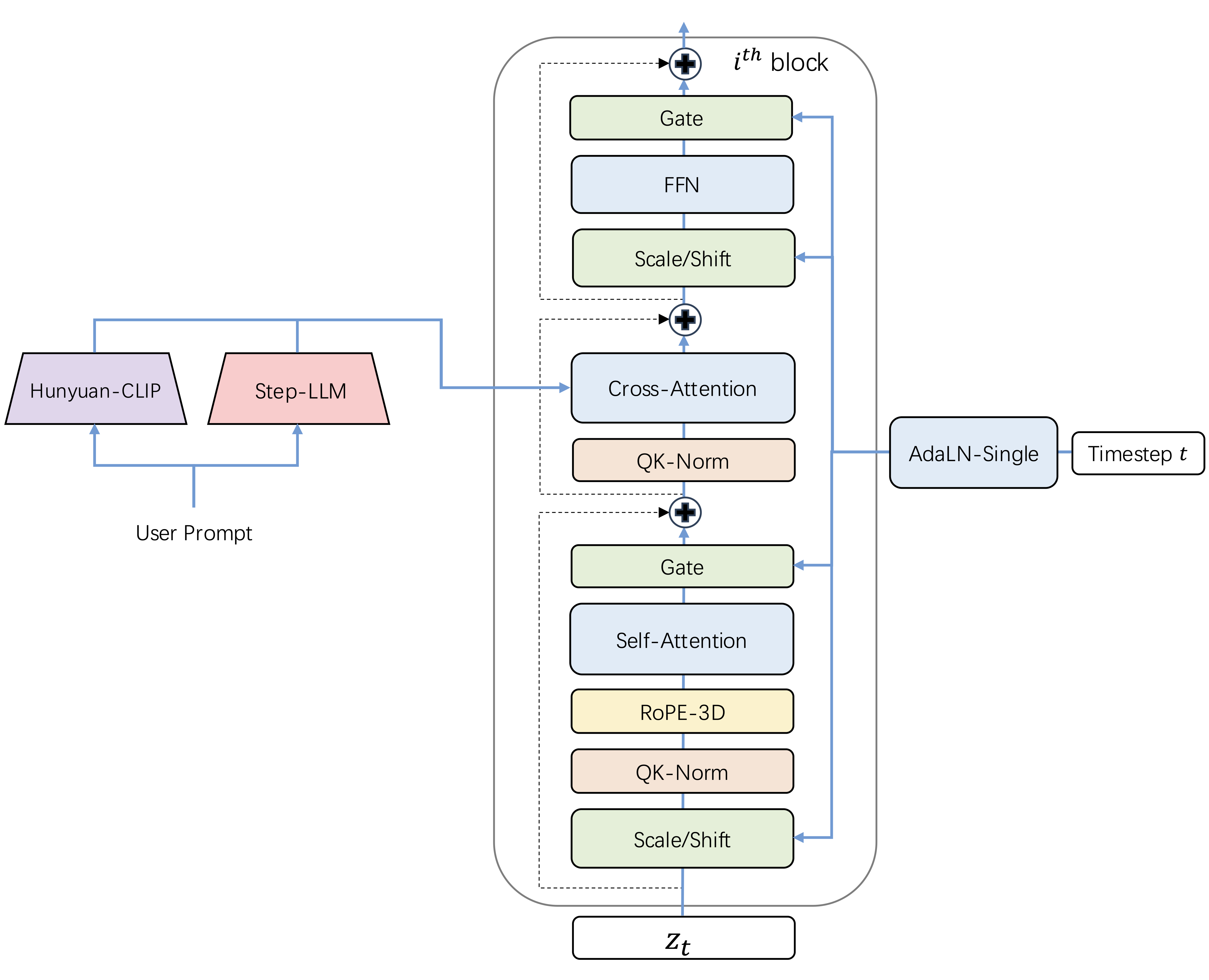}
    \caption{The model architecture of our bilingual text encoder  and DiT with 3D Attention.}
\end{figure}

\begin{table}[ht]
\centering
\resizebox{\textwidth}{!}{
\begin{tabular}{c|c|c|c|c|c|c}
\hline
Layers& Attention Heads  & Head Dim & FFN Dim & Cross-Attn Dim & Activation Function & Normalization \\
\hline
\hline
48 & 48 & 128 & 24,576 & (6,144, 1,024) & GELU-approx & RMSNorm \\
\hline
\end{tabular}
}
\caption{Hyper-parameters used in Step-Video-T2V.}
\label{hyper}
\end{table}

Step-Video-T2V is built on the DiT \cite{peebles2023scalablediffusionmodelstransformers} architecture, which consists of 30B parameters and contains 48 layers. Each layer contains 48 attention heads, with each head’s dimension set to 128. The setting of hyper-parameters used in Step-Video-T2V is outlined in Table~\ref{hyper}.

\textit{3D Full-Attention}: We employ the 3D full-attention in Step-Video-T2V instead of the spatial-temporal attention, which is more computationally efficient. This choice is driven by its theoretical upper bound for modeling both spatial and temporal information in videos, as well as its superiority in generating videos with smooth and consistent motion observed from large-scale experiments.

\textit{Cross-Attention for Text Prompt}: We introduce a cross-attention layer between the self-attention and feed-forward network (FFN) in each transformer block to incorporate text prompts. This layer enables the model to attend to textual information while processing visual features. The prompt is embedded using two distinct bilingual text encoders, Hunyuan-CLIP and Step-LLM, as described in \S\ref{textencoder}. The outputs from these two encoders are concatenated along the sequence dimension, creating the final text embedding sequence. This combined embedding is then injected into the cross-attention layer, allowing the model to generate videos conditioned on the input prompt.

\textit{Adaptive Layer Normalization (AdaLN) with Optimized Computation}: In standard DiT, each block includes an adaptive layer normalization (AdaLN) operation to embed timestep and class label information. Since the text-to-video task does not require class labels, we remove class labels from AdaLN. Furthermore, we follow \cite{chen2023pixartalphafasttrainingdiffusion} and adopt the AdaLN-Single structure to reduce the computational overhead of traditional AdaLN operations and improve overall model efficiency. In the first layer of the model, AdaLN uses an MLP block to embed timestep information. In subsequent layers, a learnable parameter is initialized to summarize the timestep embeddings, which are then used as parameters for the adaptive normalization in each block.

\textit{RoPE-3D:}
We use RoPE-3D, an extension of the traditional Rotation-based Positional Encoding (RoPE) \cite{su2023roformerenhancedtransformerrotary}, specifically designed to handle video data by accounting for temporal (frame) and spatial (height and width) dimensions. The original RoPE-1D applies a rotational transformation to positional encodings to enable flexible and continuous representation of positions in sequences of varying lengths. The rotational transformation is applied by rotating the positional encoding $\mathbf{E}_i$ at position $i$ by an angle $\theta_i = \frac{2\pi i}{\tau}$, where $\tau$ is a period controlling the rotation rate, and the resulting encoding $\mathbf{P}_i = \text{Rot}(\mathbf{E}_i, \theta_i)$ is obtained. To extend this to video data, we introduce RoPE-3D. This method splits the query and key tensors along the channel dimension, applying RoPE-1D independently to each tensor for the temporal (frame) and spatial (height and width) dimensions. The resulting encodings are then concatenated. This approach enables the model to handle video inputs with varying lengths and resolutions effectively. RoPE-3D offers several advantages, such as the ability to process videos with different frame counts and resolutions without being restricted by fixed positional encoding lengths. It improves generalization across diverse video data and effectively captures both spatial and temporal relationships within the video. By providing a continuous and flexible encoding for three-dimensional video data, RoPE-3D enhances the model’s capacity to process and generate high-quality video content.

\textit{QK-Norm:}
We use Query-Key Normalization (QK-Norm) to stabilize the self-attention mechanism. QK-Norm normalizes the dot product between the query (Q) and key (K) vectors, addressing numerical instability caused by large dot products that can lead to vanishing gradients or overly concentrated attention. This normalization ensures stable attention during training, accelerates convergence, and improves efficiency, allowing the model to focus on learning meaningful patterns. Additionally, QK-Norm helps maintain a balanced distribution of attention weights, enhancing the model's ability to capture relationships within the input sequence.

\subsubsection{Training Objective for Video and Image Generation}
We use Flow Matching in the training of Step-Video-T2V. At each training step, we begin by sampling a Gaussian noise, $X_0 \sim \mathcal{N}(0,1)$, and a random timestep $t \in [0,1]$. We then construct the model input $X_t$ as a linear interpolation between $X_0$ and $X_1$, where $X_1$ is the target sample corresponding to the noise-free input. Specifically, we define $X_t$ as:
$X_t = (1 - t) \cdot X_0 + t \cdot X_1$. The ground truth velocity $V_t$, which represents the rate of change of $X_t$ with respect to the timestep $t$, is defined as:
\begin{equation}
    V_t = \frac{dX_t}{dt} = X_1 - X_0 .
\end{equation}
In other words, $V_t$ captures the direction and magnitude of change from the initial noise $X_0$ to the target data $X_1$.
The model is then trained by minimizing the mean squared error (MSE) loss between the predicted velocity $u(X_t, y, t; \theta)$ and the true velocity $V_t$. Here, $u(X_t, y, t; \theta)$ denotes the model's predicted velocity at timestep $t$, given input $X_t$ and an optional conditioning input $y$ (e.g., a bilingual sentence). The training loss is given by:
\begin{equation}
    \text{loss} = \mathbb{E}_{t, X_0, X_1, y} \left[ \| u(X_t, y, t; \theta) - V_t \|^2 \right],
\end{equation}
where the expectation is taken over all training samples, with $t$ being the random timestep, and $X_0$, $X_1$, and $y$ drawn from the dataset. The term $\theta$ represents model parameters.
This approach ensures that the model learns to predict the instantaneous rate of change of the noisy sample $X_t$ with respect to $t$, which can later be used to reverse the diffusion process and recover data samples from noise.

\subsubsection{Inference}
During inference, we begin by sampling random noise $X_0 \sim \mathcal{N}(0, 1)$. The goal is to recover the denoised sample $X_1$ by iteratively refining the noise through an ODE-based method. For simplicity, we adopt a Gaussian solver and define a sequence of timesteps $\{ t_0, t_1, \dots, t_n \}$, where $t_0 = 0$, $t_n = 1$, and $t_0 < t_1 < \dots < t_n$. The denoising process is then carried out by integrating over these timesteps. Specifically, the denoised sample $X_1$ can be expressed as:
\begin{equation}
    X_1 = \sum_{i=0}^{n-1} u(X_{t_i}, y, t_i; \theta) \cdot (t_{i+1} - t_i),
\end{equation}
where $u(X_{t_i}, y, t_i; \theta)$ represents the predicted velocity at timestep $t_i$, given the noisy sample $X_{t_i}$ and an optional conditioning input $y$. The integral is computed over the timesteps from $t_0$ to $t_n$, with each term $u(X_{t_i}, y, t_i; \theta)$ multiplied by the corresponding timestep difference $(t_{i+1} - t_i)$.
This iterative process allows the model to gradually denoise the input sample, starting from the noise $X_0$ and progressing toward the target sample $X_1$ over the defined timesteps.

\subsection{Video-DPO}
\label{dpo}  

\begin{figure*}[t]
    \centering
    \includegraphics[width=\textwidth]{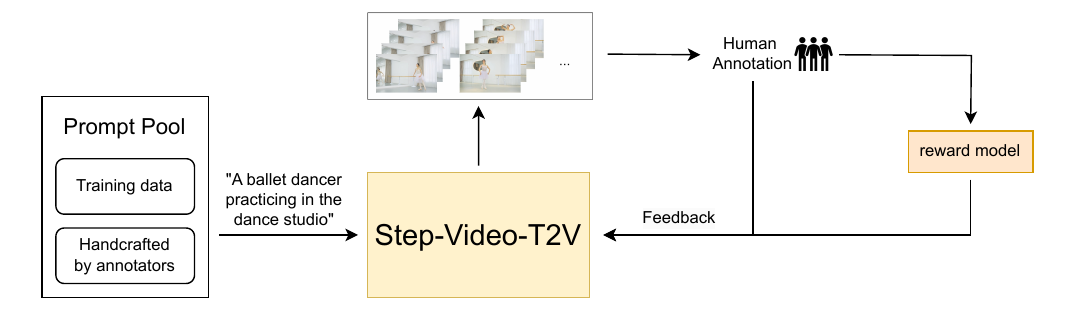}
    \caption{Overall pipeline of incorporating human feedback. }
    \label{fig:dpopipe}
\end{figure*}

The integration of human feedback has been widely validated in the domain of LLMs, particularly through methods such as Reinforcement Learning with Human Feedback (RLHF) ~\cite{ouyang2022training,christiano2017deep}, where models adjust their generated content based on human feedback. Recently, this practice has also been applied in image and video generation, yielding significant advancements. To improve the visual quality of Step-Video-T2V, we design a pipeline to introduce human feedback. 
The overall pipeline is shown in Figure~\ref{fig:dpopipe}, and details are discussed in the following.


In Step-Video-T2V, we select Direct Preference Optimization (DPO)~\cite{rafailov2024direct} as the method for incorporating human feedback. It has been proven effective across a variety of generation tasks \cite{wallace2024diffusion, yang2024using}, and the essence of the method is simple, making it both intuitive and easy to implement. Intuitively, given human preference data and non-preference data under the same conditions, the goal is to adjust the current policy (i.e., the model) to be more aligned with the generation of preferred data, while avoiding the generation of non-preferred data. To stabilize training, the reference policy (i.e., the reference model) is introduced to prevent the current policy from deviating too far from the reference policy. The policy objectvie can be formulated as:
\begin{equation}\label{eq:dpo_loss}
\mathcal{L}_{\text{DPO}} = -\mathbb{E}_{(y, x_w, x_l) \sim \mathcal{D}} \left[ 
\log \sigma \left( 
\beta \left( 
\log \frac{\pi_\theta(x_w|y)}{\pi_{\text{ref}}(x_w|y)} 
- \log \frac{\pi_\theta(x_l|y)}{\pi_{\text{ref}}(x_l|y)}
\right)
\right)
\right]
\end{equation}
where $\pi_\theta$ and $\pi_{ref}$ refers to current policy and reference policy, respectively, $x_w$ and $x_l$ are the preferred sample and non-preferred sample, and $y$ denotes the condition. 

\begin{figure*}[t]
    \centering
    \includegraphics[width=1.3\textwidth, center, trim=0 0 0 0, clip]{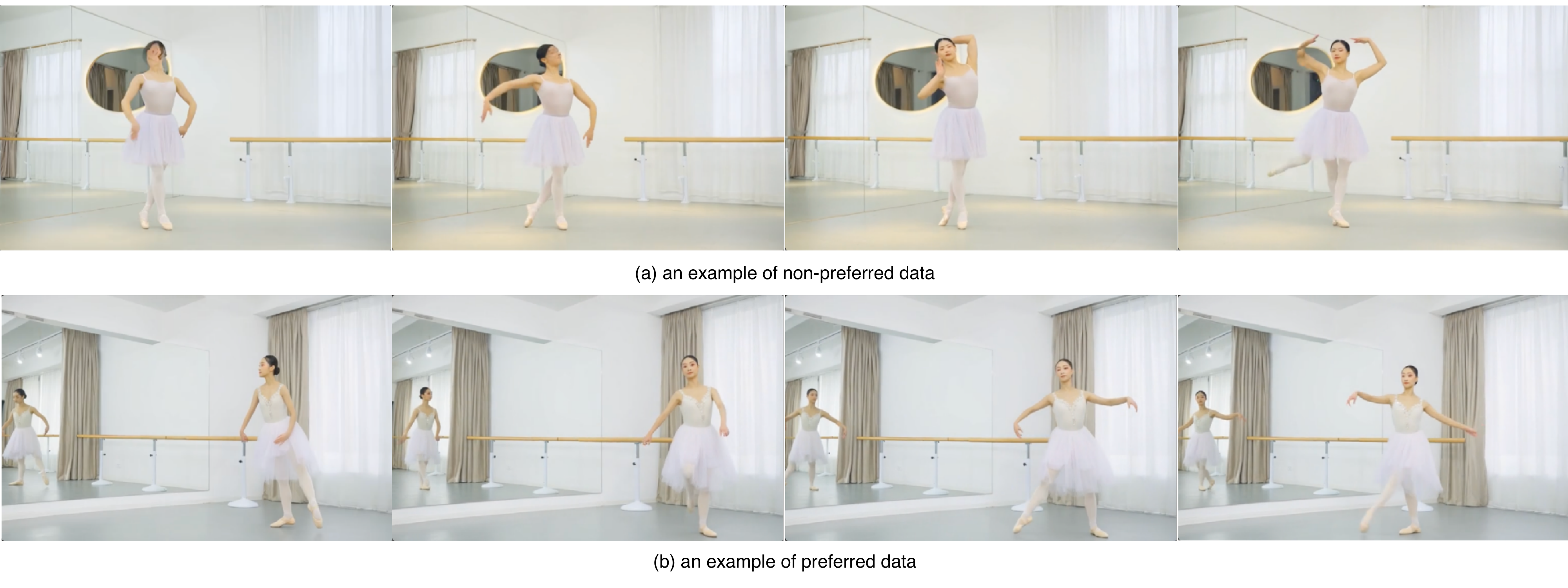}
    \caption{We generate different samples with same prompt ("A ballet dancer practicing in the dance studio" in this case), and annotate these samples as non-preferred (a) or preferred (b). }
    \label{fig:dpodata}
\end{figure*}

To collect these samples ($x_w$, $x_l$ given $y$) for training, we construct a diverse prompt set. First, we randomly select a subset of prompts from the training data to ensure prompt diversity. Second, we invite human annotators to synthesize prompts based on a carefully designed guideline that mirrors real-world user interaction patterns. And then, for each prompt, Step-Video-T2V generates multiple videos using different seeds. Human annotators rate the preference of these samples. The annotation process is monitored by quality control personnel to ensure accuracy and consistency. This process results in a set of preference and non-preference data, which serves as the foundation for model training. Two labeled examples are shown in Figure~\ref{fig:dpodata}. 

At each training step, we select a prompt and its corresponding positive and negative sample pairs described above. Each sample is generated by the model itself, ensuring smooth updates and improving overall training stability. In addition, to maintain consistency in the training data, we align the positive and negative samples by fixing the initial noise and timestep, which contributes to a more stable training process. 
Our training objective in Eqn.~\ref{eq:dpo_loss} is based on the DiffusionDPO method~\cite{wallace2024diffusion} and DPO~\cite{rafailov2024direct} but with slight modifications, extending it to the Flow Matching framework. By denoting the policy-related terms in Eqn.~\ref{eq:dpo_loss} as inside term $z$, it can be derived that:
\begin{equation}\label{eq:dpo_beta}
\frac{\partial \mathcal{L}_{\text{DPO}}}{\partial \theta} \propto -\beta (1 - \sigma(\beta z)) \cdot \frac{\partial z}{\partial \theta},
\end{equation}
which indicates large $\beta$ (\eg 5,000 in DiffusionDPO) may cause gradient explode when $z < 0$, as it amplifies gradients by $\beta$ times. As a result, gradient clipping and an extreme low learning rate (\eg 1e-8 in DiffusionDPO) are required to ensure stable training, leading to slow convergence.To address this, we reduce $\beta$ and increase the learning rate, results much faster convergence.

Human feedback effectively improves visual quality. However, we observe that the improvements saturate when the model can easily distinguish between positive and negative samples. This phenomenon may stem from the following reason: the training data used in Video-DPO is generated by earlier versions of the model. After multiple iterations of DPO, the current policy has evolved significantly (e.g., distortions are now rare) and no longer aligns with the policies from previous iterations. Consequently, updating the current policy with outdated data from earlier iterations leads to inefficient data utilization.
To address this, we propose training a reward model using human-annotated feedback data. This reward model dynamically evaluates the quality of newly generated samples during training. The reward model is periodically fine-tuned with newly annotated human feedback to maintain alignment with the evolving policy. By integrating it into the pipeline, we score and rank training data on-the-fly (on-policy), thereby improving data efficiency.

\section{Distillation for Step-Video-T2V Turbo}
Diffusion models for video generation typically require substantial computational resources during inference, often necessitating more than 50 steps of ODE integration to produce a video. Reducing the number of function evaluations (NFE) is crucial for improving inference efficiency. We demonstrate that a large-scale trained Video DiT can reduce NFE to as few as 8 steps with negligible performance degradation. This is achieved through self-distillation with a rectified flows objective and a specifically designed inference strategy.

Our base model is trained using rectified flow, and the distillation objective aims to train a 2-rectified-flow model~\citep{Liu2022FlowSA}, which facilitates more direct ODE paths during inference. As discussed by~\cite{lee2024improving}, the loss function for the 2-rectified flow can be formulated as follows:
\begin{equation}
\mathcal{L}(\theta, t) := \frac{1}{t^2} \mathbb{E}[\|v - u_\theta(x_t, t)\|_2^2] = \frac{1}{t^2} \mathbb{E}[\|x - \mathbb{E}[x|x_t]\|_2^2] + \tilde{\mathcal{L}}(\theta, t). 
\end{equation}
Since all training samples are generated by the base 1-rectified model, the irreducible loss (first term) is relatively small. The reducible error (second term) can be efficiently optimized by assigning more weight to timesteps that are more challenging. Specifically, the training loss of 2-rectified flow is large at each end of the interval \( t \in [0, 1] \) and small in the middle.

We sampled approximately 95,000 data samples using a curated distribution of SFT data prompts with 50 NFE and carefully designed positive and negative prompts to formulate a distillation dataset. We modified the timestep sampling strategy to a U-shaped distribution, specifically \( p_t(u) \propto \exp(a u) + \exp(-a u) \) on \( u \in [0, 1] \), with a larger \( a = 5 \) as the time shift required by the video model is higher.

\begin{figure}[ht]
    \centering
    \includegraphics[width=1.3\textwidth, center, trim=0 0 0 0, clip]{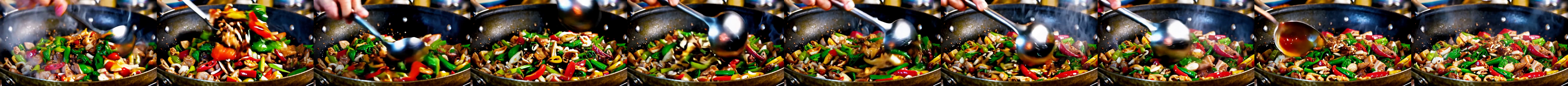}
    \includegraphics[width=1.3\textwidth, center, trim=0 0 0 0, clip]{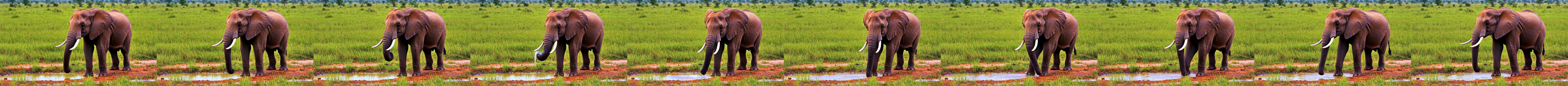}
    \includegraphics[width=1.3\textwidth, center, trim=0 0 0 0, clip]{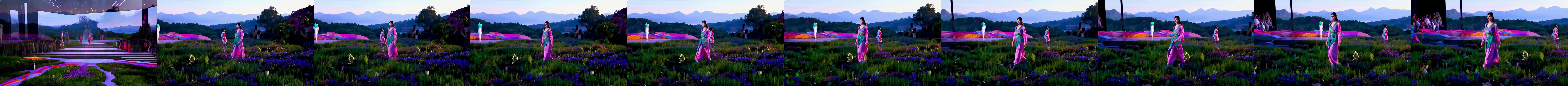}
    \includegraphics[width=1.3\textwidth, center, trim=0 0 0 0, clip]{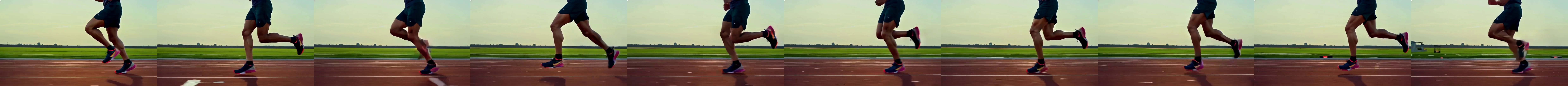}
    \includegraphics[width=1.3\textwidth, center, trim=0 0 0 0, clip]{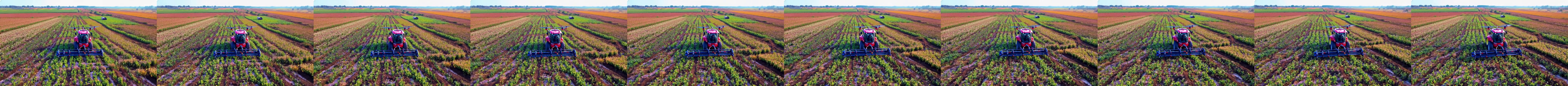}
    \includegraphics[width=1.3\textwidth, center, trim=0 0 0 0, clip]{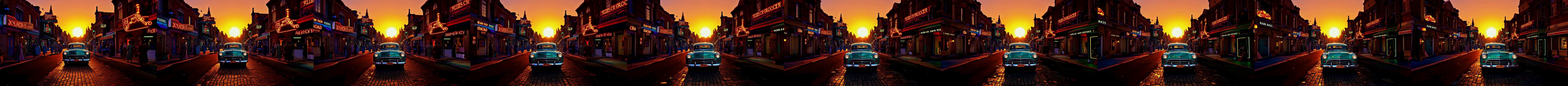}
    \caption{Generated samples with Step-Video-T2V Turbo with 10 NFE.}
    \label{fig:distillation_result}
\end{figure}

During inference, we observed that as the training progresses, the model requires more significant sampling time shifts and a lower classifier-free guidance (CFG) scale. By combining this with a linear diminishing CFG schedule as described in Eqn.~\ref{eq:cfg}, our model can achieve comparable sample quality with up to 10 times fewer steps. Figure~\ref{fig:distillation_result}, shows generated samples with 204 frames from our turbo model with 10 NFE.

\begin{equation}
\text{cfg}_t = \max(\text{cfg}_{\max} - 9t (\text{cfg}_{\max} - 1), 1) \quad \text{for} \quad 0 \leq t \leq 1
\label{eq:cfg}
\end{equation}

\section{System}

This section describes our infrastructure that facilitates the efficient and robust training of Step-Video-T2V at scale. The discussion starts with a comprehensive system overview, providing a holistic perspective of the workflow, followed by an in-depth examination of each constituent component. Furthermore, we present our insights and practical experiences gained from our training platform implementation and routine operational management.

\subsection{Overview}

\begin{figure*}[t]
    \centering
    \includegraphics[width=\textwidth]{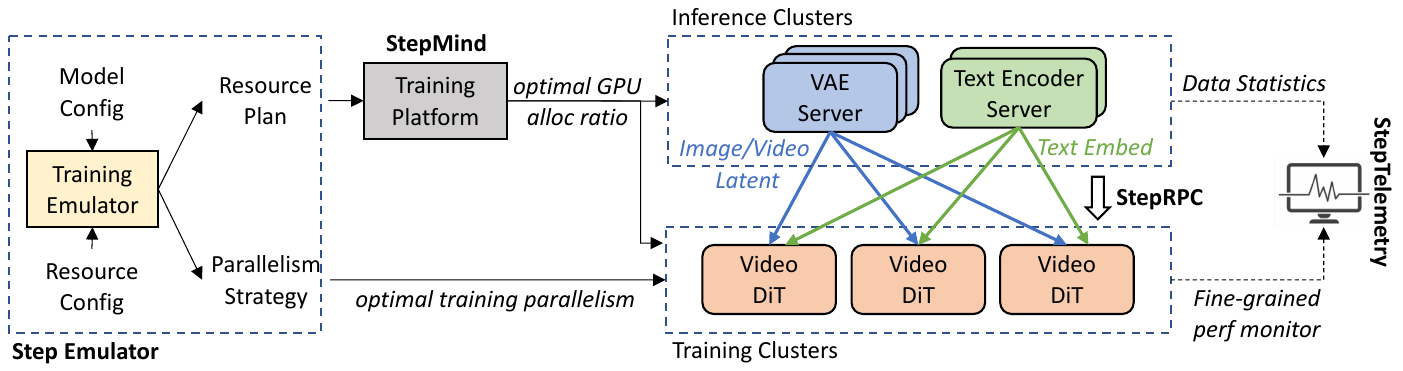}
    \caption{The workflow of Step-Video-T2V training system.}
    \label{fig:sys_overview}
\end{figure*}

Figure~\ref{fig:sys_overview} shows the overall workflow of Step-Video-T2V training system. 
The workflow comprises several stages. The offline stage, based on our in-house training emulator (Step Emulator, \S\ref{subsec:system_framework}), is specifically designed to estimate and determine optimal resource allocation and training parallelism strategies. 
This determination is achieved through systematic analysis of model architectures and resource configurations provided as input parameters. 
Next, with the theoretical optimal resource allocation plan, we deploy the training job with GPUs allocated in the training and inference clusters, respectively. The training clusters, responsible for training video DiT, uses the parallelization strategy recommended by the emulator, which has been specifically optimized to maximize the Model Flops Utilization (MFU). 
The other side with VAE and Text Encoder, runs on the inference clusters, and constantly provides the processed input data (pairs of image/video latent and text embedding) needed for DiT training. Data transmission between clusters is facilitated by StepRPC (\S\ref{subsec:system_steprpc}), our high-performance RPC framework that seamlessly integrates both TCP and RDMA protocols, enabling efficient cross-cluster communication with robust fault tolerance.

To enable systematic monitoring and analysis during large-scale training, we implement a dual-layer monitoring approach through StepTelemetry (\S\ref{subsec:system_telemetry}). This system collects detailed data statistics from inference clusters while simultaneously collecting iteration-level, fine-grained performance metrics from training clusters. The resulting telemetry data provides multidimensional insights into the training system, enabling precise identification of algorithmic patterns and systematic detection of potential performance bottlenecks across the entire infrastructure.

We have constructed a datacenter comprising thousands of NVIDIA H800 GPUs interconnected by a rail-optimized RoCEv2 fabric (1.6Tbps bandwidth per node). Nodes of the datacenter can be dynamically assigned to inference clusters or training clusters according to GPU resource requirements.
To support a single large-scale training job with thousands of GPUs spanning multiple GPU clusters concurrently, we have gained valuable insights from addressing challenges related to the training platform (StepMind) and its operational complexities. A detailed examination of these findings, including specific implementation strategies and best practices, will be presented in \S\ref{subsec:system_platform}. Through comprehensive improvements to infrastructure reliability, we have achieved \textbf{99\% effective GPU training time} over more than one month.

\subsection{Training Framework Optimizations}
\label{subsec:system_framework}

\subsubsection{Step Emulator}
\label{subsubsec:system_semu}
The large model size and extended context length of video require partitioning both the model parameters and activations/gradients across devices using multiple parallelism strategies during training, such as Tensor-parallelism (TP), Sequence-parallelism (SP), Context-parallelism~(CP), Pipeline-parallelism (PP) and Virtual Pipeline-parallelism (VPP)~\cite{pp2021,tpsp2023, scp, hcp}. 
However, the large scale of GPU cluster required for DiT training poses significant challenges in tuning and validating architecture designs and optimizations. To address this, we developed Step Emulator (SEMU), a highly accurate and efficient simulator designed to estimate resource consumption and end-to-end performance during training, under various model architecture and parallelism configurations. Specifically, to accommodate the dynamic and mixed input data for DiT training, SEMU allows customization of input data with varying frames and resolutions. SEMU helps to design model parameters,  architecture and the associated optimal parallelism strategies. It also determines the resource allocation of inference~(\textit{i.e.}, text-encoder and VAE) and training~(\textit{i.e.}, video DiT) clusters before the training actually starts.

\subsubsection{Distributed Training}
\label{subsubsec:system_dist}

\paragraph{Parallelism Strategy}
Table~\ref{tab:540p_mfu} outlines the MFU of different configurations for 540P video pre-training obtained by SEMU. As shown in the table, simply applying PP on top of TP does not achieve a high MFU. This is because PP only reduces memory usage for model parameters and gradients by about 20GB after 8-way TP, and it can only disable a small portion of activation checkpointing, given the 120GB of activation memory. While CP directly reduces activation memory, its communication cost through the NIC is comparable to the TP cost via NVLink. To reduce CP overhead, we apply head-wise CP~\cite{hcp} to the self-attention block, leveraging the MHA in the DiT model, and sequence-wise CP~\cite{scp} to the cross-attention block, due to the relatively short sequence length of $k$ and $v$ from the prompts. Despite these optimizations, the CP cost remains non-negligible, and relying solely on CP does not lead to a high MFU.

As a result, the optimal MFU is always achieved by combining TP, CP, PP, and VPP. However, for large-scale GPU cluster training, it is crucial to keep the backend framework as simple as possible for robustness and easy identification of stragglers during training. This hinders us from adopting PP since it generally lacks the necessary flexibility. As a trade-off, we adopt an 8-way tensor parallelism (TP) strategy combined with sequence parallelism (SP) and Zero1~\cite{zero2020}. This configuration results in a MFU that is marginally lower (-0.88\%) than the theoretical optimum. In practice, the actual training MFU reaches 32\%, which is slightly below the estimated value due to metric collection overhead and minor delays caused by stragglers.

\begin{table}[h]
\centering
\begin{tabular}{cccccc}
\hline
\textbf{TP}        & \textbf{CP}                 & \textbf{PP} & \textbf{VPP} & \textbf{Checkpointing (\%)} & \textbf{MFU}         \\ \hline
\multirow{2}{*}{4} & \multirow{2}{*}{1} & 2           & 24           & 93.75                   & 35.90                \\
                   &                    & 4           & 24           & 93.75                   & 35.71                \\ \hline
\multirow{7}{*}{8} & \multirow{3}{*}{1} & 1           & 1            & 83.33                   & {\underline{35.59}} \\
                   &                    & 2           & 24           & 72.91                   & 36.06                \\
                   &                    & 4           & 12           & 72.91                   & 35.76       \\ \cline{2-6} 
                   & \multirow{2}{*}{2} & 1           & 1            & 62.50                   & 31.79                \\
                   &                    & 4           & 12           & 31.25                   & 35.11                \\ \cline{2-6} 
                   & \multirow{2}{*}{3} & 1           & 1            & 31.25                   & 33.41                \\
                   &                    & 4           & 12           & 11.53                   & \textbf{36.47}                \\ \bottomrule
                   
\end{tabular}
                   
\vspace{0.1cm}
\caption{Estimated MFU from SEMU of different parallelism strategies under 540P video pre-training stage.}\label{tab:540p_mfu}
\end{table}

\paragraph{TP overlap} To minimize TP overhead, we have developed StepCCL, a proprietary collective communication library that implements advanced communication-computation overlap techniques. StepCCL directly utilizes the DMA engine for data transmission, completely bypassing the Stream Multiprocessors (SMs). This design enables simultaneous execution of StepCCL operations and GEMM computations on the same GPU, achieving true concurrency without mutual performance interference, thereby maximizing hardware utilization and computational throughput. More details can be found at Section 7 of \cite{disttrain}.

\paragraph{DP overlap}
In the first two stages~(i.e. Image and 192P video pre-training), the context length is below 10K and activation memory does not pose a limiting factor. The primary memory usage stems from model parameters, which are handled via 8-way TP. While the performance bottleneck arises from gradient reduce-scatter and parameter all-gather operations introduced by DP, which can take up more than 30\% of the training time. To mitigate this, we developed DP overlap, where the parameter all-gather is performed during the forward pass of the first micro-batch, while the gradient reduce-scatter overlaps with the backward pass of the last micro-batch. Note that in DiT training, the activation norm is typically a key metric in the training process, which registers forward hooks for monitoring. These forward hooks can slow down the kernel launch in forward process, further rendering the forward overlap of DP communication ineffective. Therefore, the effectiveness of forward overlap may vary depending on the scenario, and the decision to enable it should be made carefully on a case-by-case basis.

\subsubsection{Tailored Computation and Communication Modules}

\paragraph{VAE Computation} 
To accelerate the convolution op (the most compute-intensive) in VAE, we employ the channel-last principle that is more GPU-friendly~\cite{channel_last} than naive PyTorch implementation. Specifically, a raw PyTorch tensor uses the NCHW memory format by default, while GPU tensorcores only support NHWC format essentially, causing additional format transformation overhead that slows down overall speed. We solve this by performing format permutation at the beginning, putting the channel to the last dimension physically. We modify each op (\textit{e.g.,} Conv, GroupNorm) along the computation graph to adapt to channel-last format. Overall, we achieve up to 7x VAE encode throughput with this optimization.

\paragraph{Multi-GPUs VAE} To further reduce VAE latency and also support long, high resolution videos, using multiple GPUs is necessary to reduce computation and memory footprint on a single device. We support both Temporal and Spatial Parallel for the convolution op. As an example, for Temporal Parallel, we divide the video latent along the frame dimension and let each GPU hold only a subset of video frames. If a downstream Conv op requires cross-frame computation, we transfer the overlapped frames using all-to-all communication. The overhead is typically small ($<1\%$) compared to the computation time.

\paragraph{DiT} 
The plain RoPE implementation is inefficient due to the numerous time-consuming slice and concat operations required for building the embedding table and indexing. We developed a custom RoPE-3D kernel that replaces these indexing operations with efficient embedding computation, significantly improving performance.
The timestep modulation in the DiT model results in high activation memory usage, as the timestep is repeated across the sequence length, which is redundant since it remains the same within a single video clip. We implement a memory-efficient modulation operation where the timestep is repeated only during the forward process, and the non-repeated timestep is saved for the backward process. To further reduce memory costs, we fuse the LayerNorm op and the downstream timestep modulation op, eliminating the need for saving an intermediate output.

\subsubsection{DP Load Balance}
A critical challenge in large-scale video generation arises when processing mixed-resolution videos and images within the same global iteration.
Conventional approaches that segregate different resolutions into separate batches lead to significant FLOPs disparities across model instances, resulting in GPU under-utilization due to load imbalance.
Table~\ref{tab:vid_img_flops} outlines the FLOPs per sample of different resolutions.

\begin{table}[h]
\centering
\begin{tabular}{rr}
\toprule
Resolution (F, H, W) & TFLOPs per sample \\
\midrule
$ 204 \times 256 \times 256$ & 1,717.20 \\
$ 204 \times 192 \times 320$ & 1,592.61 \\
$ 136 \times 256 \times 256$ & 1,079.85 \\
$ 136 \times 192 \times 320$ & 1,004.89 \\
$  68 \times 256 \times 256$ &  509.31 \\
$  68 \times 192 \times 320$ &  475.87 \\
$   1 \times 256 \times 256$ &   44.99 \\
\bottomrule
\end{tabular}
\vspace{3mm}
\caption{FLOPs per sample of different resolutions.}
\label{tab:vid_img_flops}
\end{table}

\begin{figure*}[h]
    \centering
    \includegraphics[width=\textwidth]{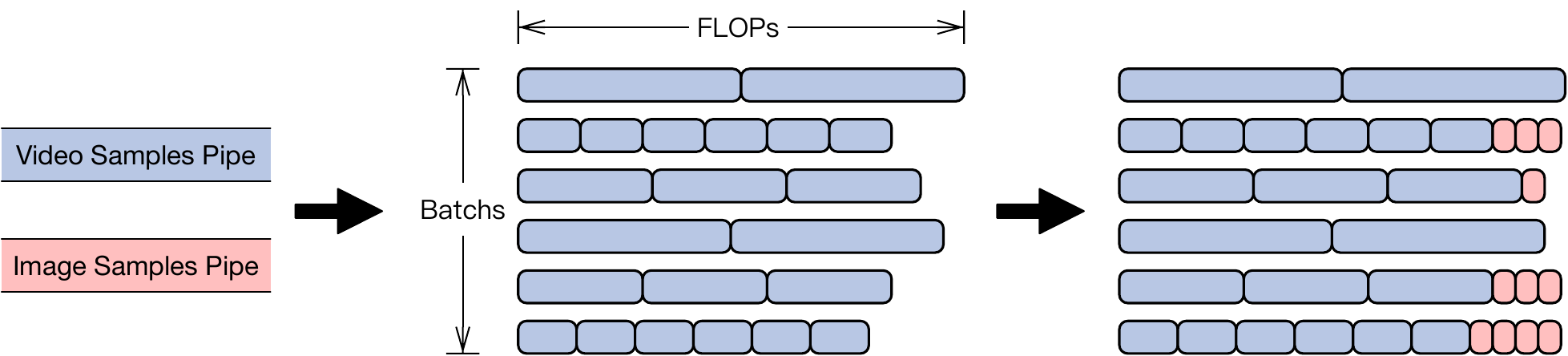}
    \caption{Load balancing with hybrid granularity.}
    \label{fig:load_balance}
\end{figure*}

To address this issue, we propose a hybrid-grained load balancing strategy that operates through two complementary stages, as illustrated in Figure~\ref{fig:load_balance}.
In the first stage, we perform \textit{coarse-grained} FLOPs alignment by adjusting batch sizes of videos with different resolutions.
For each resolution $r$, we estimate its FLOPs per sample $F_r$ and compute optimal batch sizes $B_r$ through:

\begin{equation}
B_r = \left\lfloor \frac{F_{\text{target}}}{\alpha F_r} \right\rfloor
\end{equation}

where $F_{\text{target}}$ represents the target FLOPs per batch (typically the batch of the highest resolution videos) and $\alpha$ is a normalization factor to ensure the consistency of global batch size.

The second stage addresses residual FLOPs variations through \textit{fine-grained} image padding.
Our system caches $N$ video batches and calculates required image supplements based on a predetermined video-to-image ratio $\beta$.
Using a greedy heuristic, we iteratively allocate images to the batch with the smallest current FLOPs until all supplements are distributed.

The hybrid-grained approach effectively balances computational loads while maintaining practical deployability.
Our solution requires only superficial awareness of data distribution, needing merely batch size adjustments and supplemental image padding rather than deep architectural changes.
This minimal intervention preserves the original training pipeline's integrity while introducing small memory overhead.

\subsection{StepRPC}
\label{subsec:system_steprpc}

To facilitate data transfer, we developed StepRPC, a high-performance communication framework. StepRPC leverages distributed named pipes as the core programming abstraction, enabling a large number of servers to communicate seamlessly by declaring pipes with the identical name. The spraying mode distributes data evenly across training servers. Compared to existing communication frameworks~\cite{ray,damania2023pytorchrpc,mooncacke}, StepRPC incorporates the following essential engineering optimizations.

\paragraph{Tensor-Native Communication over RDMA and TCP} In existing frameworks, tensor transfer typically entails heavy-weight serialization and deserialization overheads, amounting to tens of milliseconds. To address the inefficiency, StepRPC implements tensor-native communication that directly transfers bits within tensor memory, thereby eliminating the overheads associated with serialization and deserialization. StepRPC harnesses the power of RDMA to implement direct transfer of both GPU and CPU tensors. When RDMA-capable networks are not available, StepRPC can be seamlessly configured to utilize TCP transports. Note that TCP only supports CPU tensors. Consequently, transferring GPU tensors over TCP introduces additional overheads due to the necessity of copying memory between GPUs and CPUs. To mitigate the overheads, StepRPC proposes to overlap CudaMemcpy with TCP send and recv operations. Such optimization hides the latency of memory copying, thereby improving overall communication performance in non-RDMA environments.

\paragraph{Flexible Workload Patterns with High Resilience} To optimize GPU utilization, we leverage the same inference servers to generate data for large-scale pre-training experiments and small-scale ablation experiments simultaneously. StepRPC facilitates this via a combination of broadcasting and spraying communications. First, StepRPC broadcasts the data from inference servers to all training jobs. This ensures that each job receives the necessary data without redundant computations. Second, within an individual job, StepRPC sprays data to each training server. Though ingesting data from same inference servers, training jobs can operate independently and elastically with the help of StepRPC, meaning that jobs can begin, terminate or scale as needed without affecting the others. Meanwhile, StepRPC isolates failures across jobs, preventing cascading effects that could destabilize the entire system.

\paragraph{Enhanced Visbility for Real-Time Failure Detection and Resource Optimization} StepRPC offers comprehensive performance metrics for deep insights into the communication process. The metrics encompass critical aspects such as data counts, queuing latency and transmission cost. The enhanced visibility serves multiple purposes, empowering both operators and researchers to optimize performance and resource utilization effectively. Firstly, by monitoring the counts of produced and consumed data, StepRPC enables real-time failure detection. Discrepancies between these counts can indicate potential issues such as data loss, communication failures, or bottlenecks. This proactive approach allows operators to promptly identify and address failures. Next, researchers can leverage the metrics like queuing latency and API invoking latency to assess whether inference or training processes constitute the overall performance bottleneck.
Furthermore, armed with the metrics like rates of producing and consuming data, researchers can make informed decisions regarding GPU resource allocation for inference and training jobs.

\subsection{StepTelemetry}
\label{subsec:system_telemetry}

The lack of the observability of training framework makes analyzing it's inner state and debugging job failure difficult. Thus StepTelemetry, an observability suite for training frameworks, is introduced. This suite's goal is not only to enhance anomaly detection capabilities but also to establish a reusable pipeline for collecting, post-processing, and analyzing any training-related data.

StepTelemetry employs a simple and asynchronous data collection pipeline. It offers a Python SDK for easy integration with the training framework, supporting both batch and streaming data writes to files on local disk. An additional consumer process  is responsible for collecting, transforming, and writing data into various remote databases. StepTelemetry benefits Step-Video-T2V training in the following aspects.

\paragraph{Anomaly Detection} Common profiling tools like PyTorch Profiler and Megatron-LM Timer introduce approximately 10\% to 15\% overhead, and struggle to support collaborative analysis among multiple ranks. Instead, StepTelemetry adopts a CUDA event-based approach without any unnecessary synchronizations. This enables continuously collecting timer data of all ranks during training with almost zero overhead. By providing various data visualizations and supporting data drill-down, StepTelemetry helps pinpointing root cause in case of hardware and software failure. 
As an example, during one training session, the training efficiency fell below expectations, yet no hardware alerts were triggered. Upon analyzing the collected data, we identified that the backward propagation time for certain ranks was abnormally prolonged. Since the backward process primarily involves tensor parallelism (TP) group communication and computation, it is highly probable that the machines hosting these ranks were underperforming. After removing these machines from the training cluster, the training efficiency returned to the expected level.
\paragraph{Data Statistics} During video training, it is vital to monitor data consumption. Instead of just counting tokens, it is required to record consumed videos' metadata. The legacy approach was to dump metadata to files on local disk, and then use scripts to parse them offline, which is particularly inefficient and inconvenient. By instrumenting dataloader with StepTelemetry, the metadata is written to database, thus OLAP is enabled. Visualizations such as duplicated data filtering and data distribution monitoring based on source url is provided to researchers, which help evaluating the model.
\paragraph{Performance Optimization} StepTelemetry provides insight for performance optimization. By visualizing the time consumption of each stage within an iteration, it provides developers with a comprehensive overview, enabling them to identify and optimize performance bottlenecks in critical paths. Additionally, dataloader statistics reveal the actual throughput of the training process. Although image and video data are supplied in a mixed manner, the iteration time remained unchanged after addressing the data parallelism (DP) imbalance issue. Nevertheless, the observed increase in data throughput demonstrates a significant improvement in system efficiency.

\subsection{StepMind}
\label{subsec:system_platform}
To ensure high availability of computing resources for large-scale Step-Video-T2V training tasks, we have invested substantial efforts into developing StepMind, a distributed training platform designed for large-scale machine learning workloads. StepMind has successfully achieved an effective GPU utilization rate exceeding 99.0\% for Step-Video-T2V training, primarily through the implementation of the following key techniques.

\paragraph{Fine Grained Monitoring at Full Coverage}
To maximize distributed training efficiency, we developed a fine-grained monitoring system at full coverage that rapidly identifies faulty nodes. The monitoring system collects metrics at seconds-granularity across hardware, \eg CPU/GPU/memory/PCIe/network/storage/power/fans, and software, \eg OS stack, enabling rapid and full coverage fault detection. 
Based on our operation experiences, faulty nodes can be generally classified into two categories: 
a) Nodes with Fatal Errors (about 86.2\% of failures). 
These nodes can interrupt the training process immediately. Upon detection of these nodes, we will replace them with healthy nodes and restart the job.
In order to avoid incorrect restarts due to false alarms, we develop a multi-signal approach to ascertain whether a job requires restarting. The signals incorporated in this approach encompass RoCEv2 traffic disruption, low GPU power usage, and the cessation of updates in job training logs.
Once being identified as failed, the job will be restarted immediately, thereby reducing the time cost of unavailability resulting from node malfunctions.
b) Nodes with Non-Fatal Errors (about 13.8\% of failures).
Although these nodes do not immediately disrupt the training task, they can degrade training efficiency. Detecting such nodes is challenging, and we have developed specialized methods to identify them. These nodes are scheduled for replacement during planned maintenance, typically after a checkpoint is saved, to minimize the wasting time of computational resource.
Table~\ref{tab:training_fault_stats} shows more detailed statistics.

\begin{table}[h]
\centering
\begin{tabular}{lw{c}{3cm}w{c}{3cm}}
\toprule
\multicolumn{1}{l}{\textbf{Fault}} & \textbf{Category}  & \textbf{Count}  \\ 
\midrule
GPU\_DBE\_ERROR                         & GPU            & 3                              \\ 
GPU\_LOCKED                             & GPU            & 1                             \\ 
LINK\_DOWN                              & Network        & 1                               \\ 
NODE\_SHUTDOWN                          & Host           & 2                                 \\ 
SOFTWARE\_FAULT                         & Software       & 11                                 \\
CUDA\_OOM                               & Software       & 7                                   \\ 
NON\_FATAL                              & Hardware       & 4                      \\  \bottomrule
\end{tabular}
\vspace{0.5cm}
\caption{Over a month of Step-Video-T2V training, fatal hardware failures occurred only 7 times.}
\label{tab:training_fault_stats}
\end{table}

\paragraph{GPU Machine High Quality Ensurance}
Training GPU nodes exhibit significant quality variations, \ie their failure probabilities differ substantially. Some servers have much higher failure risks than others, necessitating the selection of the most reliable servers for large-scale training tasks to minimize the job interruptions.
We developed an innovative node quality assessment framework that systematically integrates historical alert patterns, maintenance logs, stress test results, and load test durations to generate comprehensive quality scores. When node failures occur within production resource pools, replacement units are selectively deployed from a dedicated buffer pool following a prioritized matching rule: buffer machines' quality scores must meet or exceed the operational requirements of the target resource pool's priority tier. This methodology has achieved a statistically significant reduction in failure rates for critical resource pools (\ie video pool) from an original monthly average of 7.0\% to 0.9\%. Correspondingly, the daily restart rate per 1,000 GPUs caused by hardware issues decreased to approximately 1/11 of that reported in \texttt{LLaMA3.1}~\citep{llama3}.
\begin{table}[h]
\centering
\begin{tabular}{c|w{c}{1.5cm}|w{c}{2.4cm}|w{c}{1.5cm}|w{c}{2.4cm}}
\hline
    & \multicolumn{2}{c|}{Step-Video-T2V }     & \multicolumn{2}{c}{LLaMA3.1 }   \\  \hline
Cause of Restart         & Hardware & Total Unexpected      & Hardware  & Total Unexpected       \\ \hline
Avg Daily Restarts/1k GPUs   & \textbf{0.037}    & 0.095         & \textbf{0.422}     & 0.485  \\ \hline
\end{tabular}
\caption{Restart count statistics during training for Step-Video-T2V and LLaMA3.1 .}
\label{tab:restarts_llama_compare}
\end{table}\\
Fewer restarts ultimately helped us achieve 99\% effective training time over a training period exceeding one month.
\[\text{Effective Training Time} = \frac{\text{Total Iteration Time}}{\text{Total Training Time}}\]

\paragraph{Completely Automated Server Launch Process}
When faulty machines are taken offline, they must undergo rapid repairs and meet stringent operational standards before being reintroduced into the service pool. This ensures that defective units do not negatively impact training jobs. Three key measures are implemented to achieve this:

\begin{itemize}[left=0cm]
    \item \textbf{Automated reboot-repair for transient failures.}
    A large proportion of node failures, approximately above 60\%, are transient failures. Examples include GPU DBE errors, GPU card disconnections, and network card disconnections. The transient failures can be effectively resolved by a simple restart. To speed up GPU repairing, we've created an automated system that quickly reboots servers based on the identified failure type. By integrating this reboot system with follow-up health checks and stress tests, we ensure servers can be brought online rapidly and with assured quality.
    
    \item \textbf{Comprehensive health checks via extensive diagnostic scripts.}
    We encode human expertise into reusable scripts to conduct comprehensive checks on the hardware and software configurations of GPU nodes. These checks include GPU, NIC, software driver, and firmware configurations, ensuring that servers in operation have uniform and correct hardware and software setups. In our experience, this practice prevents nodes with abnormal configurations from running training jobs, thereby reducing the likelihood of job interruptions.
    
    \item \textbf{Rigorous stress testing and admission protocols to validate performance.}
    Our comprehensive stress testing ensures each machine delivers peak performance by evaluating two key areas:
    1) Single-Machine Performance: 
    We validate GPU AI computing power (TOPS), HBM bandwidth, host-device data transfer speeds (H2D/D2H), and NVLink/PCIe connectivity between GPUs to guarantee maximum hardware capability.
    2) RDMA Network Verification
    Using PyTorch operations to simulate distributed training patterns (TP/EP/PP/DP), we test real-world network performance. Small-group testing helps swiftly identify faulty nodes, cables, or switches. Cross-GPU traffic routing through NICs enables network validation within individual machines for rapid troubleshooting.
    These tests improve node reliability and performance while preventing job failures, significantly boosting overall cluster stability and availability.

\end{itemize}

\section{Data}

\subsection{Pre-training Data}
\begin{figure*}[t]
    \centering
    \includegraphics[width=1.3\textwidth, center, trim=0 0 0 0, clip]{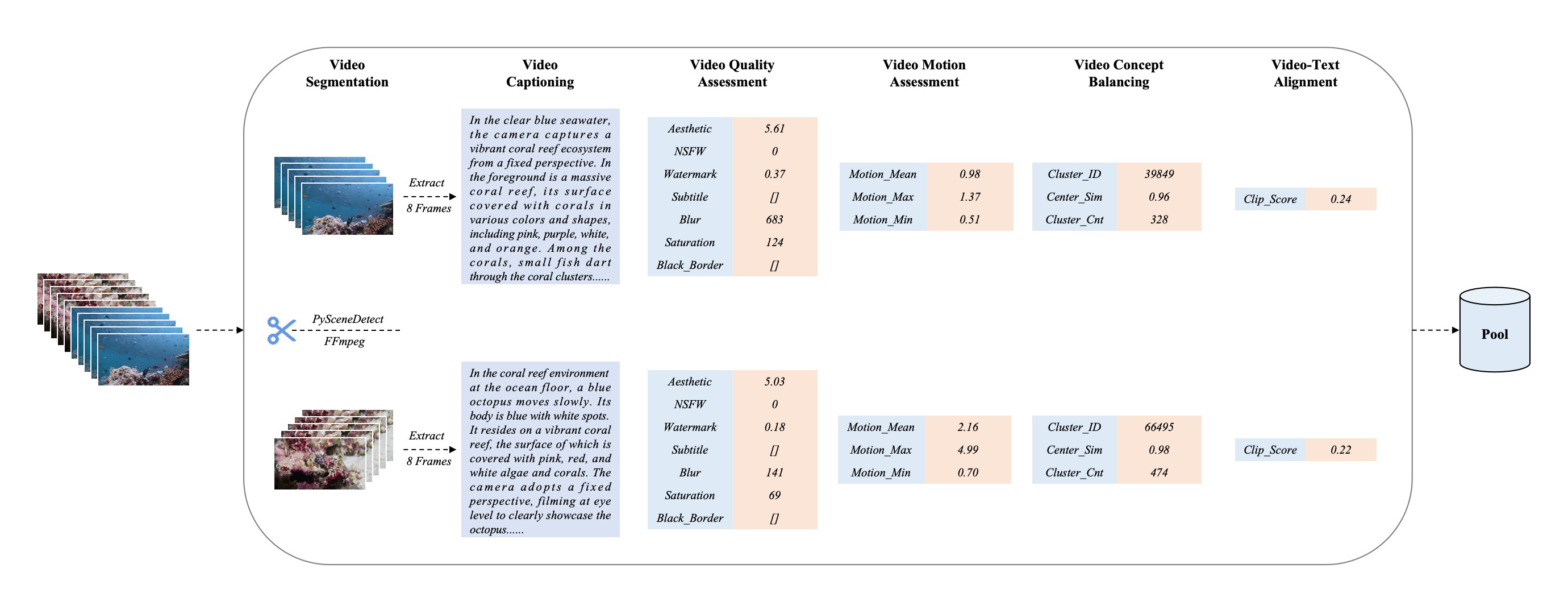}
    \caption{The pipeline of Step-Video-T2V data process.}
    \label{fig:data_pipeline}
\end{figure*}

We constructed a large-scale video dataset comprising 2B video-text pairs and 3.8B image-text pairs. Leveraging a comprehensive data pipeline, we transformed raw videos into high-quality video-text pairs suitable for model pre-training. As illustrated in Figure~\ref{fig:data_pipeline}, our pipeline consists of several key stages: Video Segmentation, Video Quality Assessment, Video Motion Assessment, Video Captioning, Video Concept Balancing and Video-Text Alignment. Each stage plays a crucial role in constructing the dataset, and we describe them in detail below.

\paragraph{Video Segmentation} 
We began by processing raw videos using the \textit{AdaptiveDetector} function in the \texttt{PySceneDetect}~\citep{pyscenedetect} toolkit to dentify scene changes and use FFmpeg~\citep{ffmpeg} to split them into single-shot clips. 
We adjusted the splitting process for high-resolution videos not encoded with \texttt{libx264} to include necessary reference frames—specifically by properly setting the crop start time in \texttt{FFmpeg}; this prevented visual artifacts or glitches in the output video.
We also removed the first three frames and the last three frames of each clip, following practices similar to Panda70M \cite{chen2024panda} and Movie Gen Video \cite{polyak2024moviegencastmedia}. Excluding these frames eliminates unstable camera movements or transition effects often present at the beginnings and endings of videos.

\paragraph{Video Quality Assessment}

To construct a refined dataset optimized for model training on high-quality, we systematically evaluated and filtered video clips by assigning multiple Quality Assessment tags based on specific criteria. We uniformly sampled eight frames from each clip to compute these tags, providing a consistent and comprehensive assessment of each video.

\begin{itemize}[left=0cm]
\item \textbf{Aesthetic Score}: We used the public LAION CLIP-based aesthetic predictor~\cite{schuhmann2022laion} to predict the aesthetic scores of eight frames from each clip and calculated their average.

\item \textbf{NSFW Score}: We employed the public LAION CLIP-based NSFW detector~\cite{laion2021nsfw}, a lightweight two-class classifier using CLIP ViT-L/14 embeddings, to identify content inappropriate for safe work environments.

\item \textbf{Watermark Detection}: Employing an EfficientNet image classification model~\cite{tan2019efficientnet}, we detected the presence of watermarks within the videos.

\item \textbf{Subtitle Detection}: Utilizing PaddleOCR~\cite{paddleocr}, we recognized and localized text within video frames, identifying clips with excessive on-screen text or captions.

\item \textbf{Saturation Score}: 
We assessed color saturation by converting video frames from BGR to HSV color space and extracting the saturation channel, using OpenCV \cite{opencv_library}. We computed statistical measures—including mean, maximum, and minimum saturation values—across the frames.

\item \textbf{Blur Score}: 
We detect blurriness by applying the variance of the Laplacian method~\cite{pech2000diatom} to measure the sharpness of each frame. Low variance values indicate blurriness caused by camera shake or lack of clarity.

\item \textbf{Black Border Detection}: We use \texttt{FFmpeg} to detect black borders in frames and record their dimensions to facilitate cropping, ensuring that the model trains on content free of distracting edges.
\end{itemize}

\paragraph{Video Motion Assessment}

Recognizing that motion content is crucial for representing dynamic scenes and ensuring effective model training, we calculate the motion score by averaging the mean magnitudes of the optical flow \cite{opencv_library} between pairs of resized grayscale frames, using the Farneback algorithm. We introduced three evaluative tags centered around motion scores:

\begin{itemize}[left=0cm]
    \item \textbf{Motion\_Mean}: The average motion magnitude across all frames in the clip, indicating the general level of motion. This score helps us identify clips with appropriate motion; clips with extremely low \texttt{Motion\_Mean} values suggest static or slow motion scenes that may not effectively contribute to training models focused on dynamic content.

    \item \textbf{Motion\_Max}: The maximum motion magnitude observed in the clip, highlighting instances of extreme motion or motion distortion. High \texttt{Motion\_Max} values may indicate the presence of frames with excessive or jittery motion.

    \item \textbf{Motion\_Min}: The minimum motion magnitude in the clip, identifying clips with minimal motion. Clips with very low \texttt{Motion\_Min} may contain idle frames or abrupt pauses, which could be undesirable for training purposes.
\end{itemize}

\paragraph{Video Captioning} 
Recent studies~\citep{openaisora, betker2023improving} have highlighted that both precision and richness of captions are crucial in enhancing the prompt-following ability and output quality of generative models. 
 
Motivated by this, we introduced three types of caption labeling into our video captioning process by employing an in-house Vision Language Model (VLM) designed to generate both short and dense captions for video clips.
\begin{itemize}[left=0cm]
    \item \textbf{Short Caption}: The short caption provides a concise description, focusing solely on the main subject and action, closely mirroring real user prompts.

    \item \textbf{Dense Caption}: The dense caption integrates key elements, emphasizing the main subject, events, environmental and visual aspects, video type and style, as well as camera shots and movements. To refine camera movements, we manually collected annotated data and performed SFT on our in-house VLM, incorporating common camera movements and shooting angles.
    
    \item \textbf{Original Title}: We also included a variety of caption styles by incorporating a portion of the original titles from the raw videos, adding diversity to the captions.
    
\end{itemize}

\paragraph{Video Concept Balancing}
To address category imbalances and facilitate deduplication in our dataset, we computed embeddings for all video clips using an internal VideoCLIP model and applied K-means clustering \cite{macqueen1967some} to group them into over 120,000 clusters, each representing a specific concept or category. By leveraging the cluster size and the distance to centroid tags, we balanced the dataset by filtering out clips that were outliers within their respective clusters. As part of this process, we added two new tags to each clip:

\begin{itemize}[left=0cm] 
    \item \textbf{Cluster\_Cnt}: The total number of clips in the cluster to which the clip belongs.

    \item \textbf{Center\_Sim}: The cosine distance between the clip's embedding and the cluster center.
\end{itemize}

\paragraph{Video-Text Alignment}

Recognizing that accurate alignment between video content and textual descriptions is essential to generate high-quality output and effective data filtering, we compute a \textbf{CLIP Score} to measure video-text alignment. This score assesses how well the captions align with the visual content of the video clips.

\begin{itemize}[left=0cm] 
\item \textbf{CLIP Score}: We begin by uniformly sampling eight frames from the given video clip. Using the CLIP model~\cite{yang2022chineseclip}, we then extract image embeddings for these frames and a text embedding for the video caption. The \texttt{CLIP Score} is computed by averaging the cosine similarities between each frame embedding and the caption embedding.

\end{itemize}

\subsection{Post-training Data}

For SFT in post-training, we curate a high-quality video dataset that captures good motion, realism, aesthetics, a broad range of concepts, and accurate captions. Inspired by \cite{dai2023emu, polyak2024moviegencastmedia, kong2024hunyuanvideo}, we utilize both automated and manual filtering techniques:

\begin{itemize}[left=0cm]

\item \textbf{Filtering by Video Assessment Scores}: Using video assessment scores and heuristic rules, we filter the entire dataset to a subset of 30M videos, significantly improving its overall quality.

\item \textbf{Filtering by Video Categories}: For videos within the same cluster, we use the "Distance to Centroid" values to remove those whose distance from the centroid exceeds a predefined threshold. This ensures that the resulting video subset contains a sufficient number of videos for each cluster while maintaining diversity within the subset.

\item \textbf{Labeling by Human Annotators}: In the final stage, human evaluators assess each video for clarity, aesthetics, appropriate motion, smooth scene transitions, and the absence of watermarks or subtitles. Captions are also manually refined to ensure accuracy and include essential details such as camera movements, subjects, actions, backgrounds, and lighting.

\end{itemize}

\section{Training Strategy}
\begin{table}[h!]
\centering
\resizebox{\textwidth}{!}{
\begin{tabular}{ccccccc}
\hline training stage & dataset & bs/node & learning rate & \#iters & \#seen samples \\
\hline 
\hline
    \multirow{3}{*}{Step-1: T2I Pre-training (256px)} & $\mathcal{O}(1) \mathrm{B}$ images & 40 & 1e-4 & 53k & 0.8B \\
     & $\mathcal{O}(1) \mathrm{B}$ images & 40 & 1e-4 & 200k & 3B \\
     \cline{2-6}
     & \textbf{Total} &  &  &  \textbf{253k} & \textbf{3.8B} \\
\hline 
\hline
    \multirow{4}{*}{Step-2: T2VI Pre-training (192px)} & $\mathcal{O}(1) \mathrm{B}$ video clips & 4 & 6e-5 & 171k & 256M\\
    & $\mathcal{O}(100) \mathrm{M}$ video clips & 4 & 6e-5 & 101k & 151M \\
    & $\mathcal{O}(100) \mathrm{M}$ video clips & 4 & 6e-5 & 158k & 237M \\
    \cline{2-6}
    & \textbf{Total} &  &  &  \textbf{430k} & \textbf{644M} \\
\hline
\hline
    \multirow{4}{*}{Step-2: T2VI Pre-training (540px)} & $\mathcal{O}(100) \mathrm{M}$ video clips & 2 & 2e-5 & 23k & 17.3M\\
    & $\mathcal{O}(10) \mathrm{M}$ video clips & 2 & 1e-5 & 17k & 8.5M \\
    & $\mathcal{O}(1) \mathrm{M}$ video clips & 1 & 1e-5 & 6k & 1.5M \\
    \cline{2-6}
    & \textbf{Total} &  &  &  \textbf{46k} & \textbf{27.3M} \\
\hline
\end{tabular}
}
\caption{Pre-training details of Step-Video-T2V. 256px, 192px, and 540px denote resolutions of 256x256, 192x320, and 544x992, respectively.}
\label{trainingrecipe}
\end{table}

\begin{figure}[h] 
    \centering
    \includegraphics[width=0.5\textwidth]{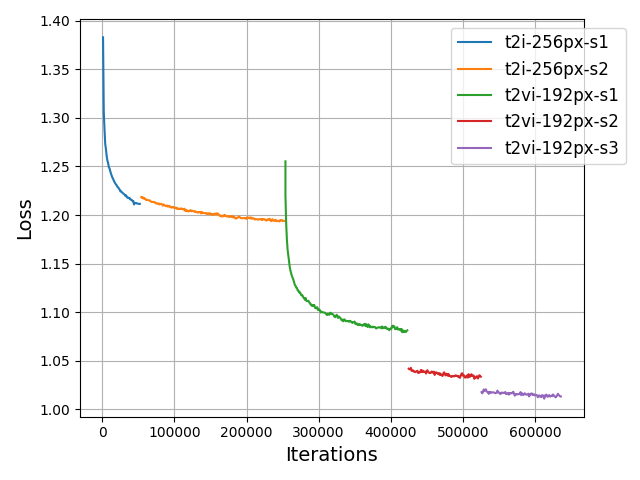}  
    \caption{Training curve of different training stages, where $s_{i}$ denotes the $i^{th}$ dataset used in the corresponding stage.} 
    \label{fig:training curve}  
\end{figure}

A cascaded training strategy is employed in Step-Video-T2V, which mainly includes four steps: text-to-image (T2I) pre-training, text-to-video/image (T2VI) pre-training, text-to-video (T2V) fine-tuning, and direct preference optimization (DPO) training. The pre-training recipe is summarized in Table~\ref{trainingrecipe}.

\paragraph{Step-1: T2I Pre-training} In the initial step, we begin by training Step-Video-T2V with a T2I pre-training approach from scratch. We intentionally avoid starting with T2V pre-training directly, as doing so will significantly slow down model convergence. This conclusion stems from our early experiments with the T2V pre-training from scratch on the 4B model, where we observed that the model struggled to learn new concepts and was much slower to converge. By first focusing on T2I, the model can establish a solid foundation in understanding visual concepts, which can later be expanded to handle temporal dynamics in the T2V phase.

\paragraph{Step-2: T2VI Pre-training} After acquiring spatial knowledge from T2I pre-training in Step-1, Step-Video-T2V progresses to a T2VI joint training stage, where both T2I and T2V are incorporated. This step is further divided into two stages. In the first stage, we pre-train Step-Video-T2V using low-resolution (192x320, 192P) videos, allowing the model to primarily focus on learning motion-related knowledge rather than fine details. In the second stage, we increase the video resolution to 544x992 (540P) and continue pre-training to enable the model to learn more intricate details. We observed that during the first stage, the model concentrates on learning motion, while in the second stage, it shifts its focus more toward learning fine details. Based on these observations, we allocate more computational resources to the first stage in Step-2 to better capture motion knowledge.

\paragraph{Step-3: T2V Fine-tuning} Due to the diversity in pre-training video data across different domains and qualities, using a pre-trained checkpoint usually introduces artifacts and varying styles in the generated videos. To mitigate these issues, we continue the training pipeline with a T2V fine-tuning step. In this stage, we use a small number of text-video pairs and remove T2I, allowing the model to fine-tune and adapt specifically to text-to-video generation.

Similar to Movie Gen Video, we found that averaging models fine-tuned with different SFT datasets improves the quality and stability of the generated videos, outperforming the Exponential Moving Average (EMA) method. Even averaging checkpoints from the same data source enhances stability and reduces distortions. Additionally, we select model checkpoints based on the period after the gradient norm peaks, ensuring both the gradient norm and loss have decreased for improved stability.

\paragraph{Step-4: DPO Training}
As described in \S\ref{dpo}, video-based DPO training is employed to enhance the visual quality of the generated videos and ensure better alignment with user prompts.

\paragraph{Hierarchical Data Filtering}
\begin{figure*}[t]
    \centering
    \includegraphics[width=1.3\textwidth, center, trim=0 0 0 0, clip]{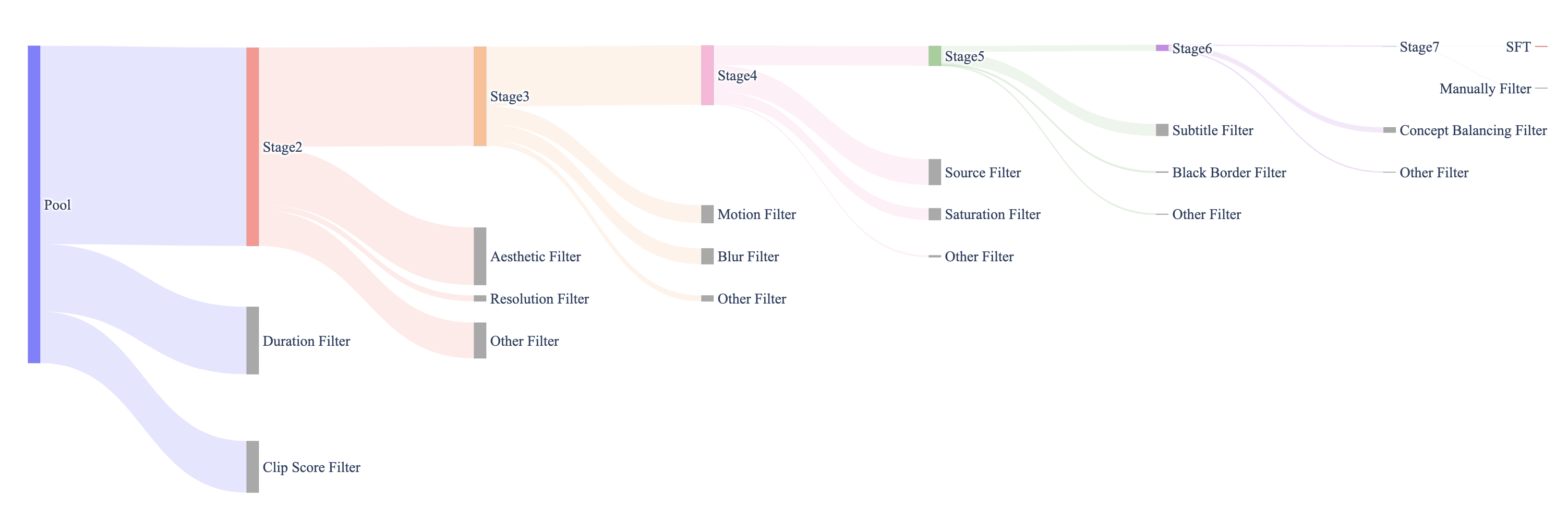}
    \caption{Hierarchical data filtering for pre-training and post-training.}
    \label{fig:data_filter}
\end{figure*}

We apply a series of filters to the data, progressively increasing their thresholds to create six pre-training subsets for Step-2: T2VI Pre-training, as shown in Table~\ref{trainingrecipe}. The final SFT dataset is then constructed through manual filtering. Figure~\ref{fig:data_filter} illustrates the key filters applied at each stage, with gray bars representing the data removed by each filter, and colored bars indicating the remaining data at each stage.

\paragraph{Observations from Pre-training Curve}
During pre-training, we observe a notable reduction in loss, which correlates with the improved quality of the training data, as illustrated in Figure \ref{fig:training curve}.

Additionally, a sudden drop in loss occurs as the quality of the training dataset improves. This improvement is not directly driven by supervision through a loss function during model training, but rather follows human intuition (e.g., filtering via CLIP scores, aesthetic scores, etc.). While the flow matching algorithm does not impose strict requirements on the distribution of the model’s input data, adjusting the training data to reflect what is considered higher-quality by humans results in a significant, stepwise reduction in training loss. This suggests that, to some extent, the model’s learning process may emulate human cognitive patterns.

\paragraph{Bucketization for Variable Duration and Size}

To accommodate varying video lengths and aspect ratios during training, we employed variable-length and variable-resolution strategies~\cite{chen2023pixartalphafasttrainingdiffusion, opensora}. We defined four length buckets (1, 68, 136, and 204 frames) and dynamically adjusted the number of latent frames based on the video length. Additionally, we grouped videos into three aspect ratio buckets—landscape, portrait, and square—according to the closest height-to-width ratio.

\section{Experiments}

\subsection{Benchmark and Metric}
We build \textbf{Step-Video-T2V-Eval}, a new benchmark for assessing the quality of text-to-video models. This benchmark consists of 128 Chinese prompts sourced from real users and is designed to evaluate the quality of generated videos across 11 categories, including Sports, Food, Scenery, Animals, Festivals, Combined Concepts, Surreal, People, 3D Animation, Cinematography, and Style.

Additionally, we propose two human evaluation metrics based on Step-Video-T2V-Eval, which can be used to compare the performance of Step-Video-T2V with that of a target model:
\begin{itemize}[left=0cm] 
\item \textbf{Metric-1} compares Step-Video-T2V with a target model by having each human annotator assign a Win/Tie/Loss label to each generated video pair from the two models for the same prompt, with the model names masked. A "Win" means Step-Video-T2V performs better than the target model, a "Loss" means it performs worse, and a "Tie" indicates the models have similar quality.
\item \textbf{Metric-2} assigns four scores to each generated video to measure its quality across the following 4 dimensions: (1) instruction following, (2) motion smoothness, (3) physical plausibility, and (4) aesthetic appeal. 
The two models are then compared based on their labeled scores. 
\end{itemize}

The criteria for scoring each dimension in Metric-2 are outlined below:
\begin{itemize}[left=0cm] 
\item \textbf{Instruction Following.} 
\textbf{Score=5}: The video is fully consistent with the prompt, with all elements and details generated accurately, and the expression of complex scenarios is flawless. 
\textbf{Score=4}: The video is generally consistent with the prompt, but there are slight discrepancies in some minor details. 
\textbf{Score=3}: The video mostly meets the prompt’s requirements, but there are noticeable deviations in several details or core content. 
\textbf{Score=2}: The video is clearly inconsistent with the prompt, with significant detail omissions or overall deviations. 
\textbf{Score=1}: The video is completely inconsistent with the prompt, with major scenes or subjects completely incorrect.

\item \textbf{Motion Smoothness.}
\textbf{Score=5}: The motion is smooth and natural, with all movements and transitions flowing seamlessly.
\textbf{Score=4}: The motion is generally smooth, but there are occasional slight unnatural movements in certain scenes.
\textbf{Score=3}: The motion has slight unnatural or stuttering elements, but it doesn’t affect overall understanding.
\textbf{Score=2}: The motion is unnatural or disconnected, with noticeable stuttering.
\textbf{Score=1}: The motion is very unnatural, with frequent stuttering, making it difficult to understand.

\item \textbf{Physical Plausibility.}
\textbf{Score=5}: All object interactions and movements adhere to real-world physical laws, with accurate lighting, shadow, and collision effects, and smooth motion.
\textbf{Score=4}: Most of the physical behavior is reasonable, with occasional minor unnatural collisions or lighting issues, but they don’t affect the overall effect.
\textbf{Score=3}: Several instances of object motion, lighting, or interactions conflict with physical logic, but the main actions still have a degree of coherence.
\textbf{Score=2}: The physical behavior is unrealistic, with lighting or object interactions violating physical laws, making the scene appear unnatural.
\textbf{Score=1}: The physical behavior is completely incorrect, with severe distortion in object interactions or lighting, making the scene difficult to understand.

\item \paragraph{Aesthetic Appeal.}
\textbf{Score=5}: Highly captivating, deeply moving, with significant artistic value and visual appeal.
\textbf{Score=4}: Pleasant and engaging, effectively capturing the audience’s attention with good visual value.
\textbf{Score=3}: Somewhat appealing, but overall performance is mediocre and doesn’t leave a lasting impression.
\textbf{Score=2}: Average, lacking in appeal, and may cause the audience to lose interest.
\textbf{Score=1}: Unpleasant, lacking in appeal, and the overall effect is disappointing.
\end{itemize}

\subsection{Comparisons to Open-source Model}
We first compare Step-Video-T2V with HunyuanVideo on Step-Video-T2V-Eval.

\begin{table}[ht]\scriptsize
\centering
\begin{tabular}{c|c|c|c}
\hline
Step-Video-T2V vs. HunyuanVideo (Win-Tie-Loss) & Annotator-1  & Annotator-2 & Annotator-3 \\
\hline
\hline
Overall & \cellcolor{green!20}{59-22-47} & \cellcolor{green!20}{46-47-35} & \cellcolor{green!20}{54-41-33} \\
\hline
\hline
Sports & \cellcolor{green!20}{6-3-3} & \cellcolor{green!20}{5-5-2} & \cellcolor{green!20}{6-6-0} \\
\hline
Food & \cellcolor{green!20}{5-2-4} & \cellcolor{green!20}{5-4-2} & 3-7-1 \\
\hline
Scenery & \cellcolor{green!20}{5-3-4} & 2-9-1 & \cellcolor{green!20}{7-1-4} \\
\hline
Animals & \cellcolor{yellow!20}{6-0-6} & \cellcolor{yellow!20}{3-6-3} & 2-7-3 \\
\hline
Festivals & \cellcolor{green!20}{4-4-3} & \cellcolor{green!20}{5-2-4} & \cellcolor{green!20}{4-5-2} \\
\hline
Combined Concepts & \cellcolor{yellow!20}{5-2-5} & \cellcolor{green!20}{6-3-3} & \cellcolor{green!20}{8-1-3} \\
\hline
Surreal & 4-2-5 & \cellcolor{green!20}{5-2-4} & \cellcolor{green!20}{6-2-3} \\
\hline
People & \cellcolor{green!20}{6-2-4} & 3-4-5 & \cellcolor{yellow!20}{5-2-5} \\
\hline
3D Animation & \cellcolor{green!20}{7-1-4} & \cellcolor{green!20}{4-5-3} & \cellcolor{green!20}{6-3-3} \\
\hline
Cinematography & \cellcolor{yellow!20}{5-1-5} & 2-5-4 & 1-4-6 \\
\hline
Style & \cellcolor{green!20}{6-2-4} & \cellcolor{green!20}{6-2-4} & \cellcolor{green!20}{6-3-3} \\
\hline
\end{tabular}
\caption{Comparison with HunyuanVideo using Metric-1.}
\label{ranking}
\end{table}

\begin{table}[ht]\scriptsize
\centering
\resizebox{\textwidth}{!}{
\begin{tabular}{c|c|c|c|c}
\hline
Step-Video-T2V vs. HunyuanVideo & Instruction Following & Motion Smoothness & Physical Plausibility & Aesthetic Appeal \\
\hline
\hline
Overall & \cellcolor{green!20}{1,273-1,221} & \cellcolor{green!20}{1,407-1,327} & \cellcolor{green!20}{1,417-1,238} & \cellcolor{green!20}{1,312-1,238} \\
\hline
\hline
Sports & \cellcolor{green!20}{130-111} & \cellcolor{green!20}{120-104} & \cellcolor{green!20}{113-99} & \cellcolor{green!20}{110-98} \\
\hline
Food & 85-92 & \cellcolor{green!20}{110-97} & \cellcolor{green!20}{107-93} & \cellcolor{green!20}{111-90} \\
\hline
Scenery & \cellcolor{green!20}{130-129} & \cellcolor{green!20}{139-126} & \cellcolor{green!20}{134-120} & \cellcolor{green!20}{125-122} \\
\hline
Animals & 104-106 & \cellcolor{green!20}{123-114} & \cellcolor{green!20}{110-107} & 99-108 \\
\hline
Festivals & \cellcolor{green!20}{102-91} & \cellcolor{green!20}{110-102} & \cellcolor{green!20}{97-90} & \cellcolor{green!20}{103-94} \\
\hline
Combined Concepts & \cellcolor{green!20}{132-115} & \cellcolor{green!20}{139-136} & \cellcolor{green!20}{139-135} & \cellcolor{green!20}{118-115} \\
\hline
Surreal & 99-101 & 138-139 & \cellcolor{green!20}{135-134} & 125-126 \\
\hline
People & 115-117 & \cellcolor{yellow!20}{129-129} & 148-150 & \cellcolor{green!20}{115-112} \\
\hline
3D Animation & \cellcolor{green!20}{113-109} & \cellcolor{green!20}{137-133} & \cellcolor{green!20}{149-146} & \cellcolor{green!20}{139-135} \\
\hline
Cinematography & \cellcolor{green!20}{121-117} & 121-122 & 132-133 & \cellcolor{green!20}{116-115} \\
\hline
Style & \cellcolor{green!20}{142-133} & \cellcolor{green!20}{141-125} & \cellcolor{green!20}{153-134} & \cellcolor{green!20}{151-123} \\
\hline
\end{tabular}
}
\caption{Comparison with HunyuanVideo using Metric-2. We invited three human annotators to evaluate each video. For each category and evaluation dimension, we aggregated the scores given by all annotators across all prompts within the category for that dimension.}
\label{ranking-hy2}
\end{table}

From Table~\ref{ranking} and Table~\ref{ranking-hy2} we got three observations.

First, Step-Video-T2V demonstrates state-of-the-art performance as the strongest open-source text-to-video generation model to date. This success is attributed to multiple factors, including the model’s structural design and its pre-training and post-training strategies.
Second, in some categories like Animals, Step-Video-T2V performs worse than HunyuanVideo, as shown in Table~\ref{ranking}. This is primarily due to aesthetic issues, as verified by the Aesthetic Appeal score in Table~\ref{ranking-hy2}.
Third, Video-VAE achieves compression ratios of 16x16 spatial and 8x temporal, compared to HunyuanVideo’s 8x8 spatial and 4x temporal compression. This higher compression rate enables Step-Video-T2V to generate videos up to 204 frames, nearly double the 129-frame maximum of HunyuanVideo.

\subsection{Comparisons to Commercial Model}
We then compare Step-Video-T2V with two leading text-to-video engines in China, T2VTopA (2025-02-10 version) and T2VTopB (2025-02-10 version), on Step-Video-T2V-Eval.

\begin{table}[ht]\scriptsize
\centering
\begin{tabular}{c|c|c|c}
\hline
Step-Video-T2V vs. T2VTopA (Win-Tie-Loss) & Annotator-1  & Annotator-2 & Annotator-3 \\
\hline
\hline
Overall & 44-13-69 & 41-13-72 & 46-25-55 \\
\hline
\hline
Sports & \cellcolor{green!20}{6-2-4} & \cellcolor{green!20}{7-0-5} & \cellcolor{green!20}{7-3-2} \\
\hline
Food & \cellcolor{green!20}{5-2-4} & \cellcolor{green!20}{6-1-4} & 4-2-5 \\
\hline
Scenery & 1-0-10 & 4-0-7 & 1-2-8 \\
\hline
Animals & 1-3-8 & 1-3-8 & 3-1-8 \\
\hline
Festivals & \cellcolor{green!20}{6-2-3} & \cellcolor{green!20}{7-2-2} & \cellcolor{green!20}{5-3-3} \\
\hline
Combined Concepts & 2-0-10 & 1-3-8 & \cellcolor{green!20}{8-0-4} \\
\hline
Surreal & 4-1-6 & 3-2-6 & 4-2-5 \\
\hline
People & 2-1-8 & 2-1-8 & \cellcolor{green!20}{6-1-4} \\
\hline
3D Animation & \cellcolor{yellow!20}{6-0-6} & 3-0-9 & \cellcolor{green!20}{5-3-4} \\
\hline
Cinematography & \cellcolor{yellow!20}{5-1-5} & 4-1-6 & 1-3-7 \\
\hline
Style & \cellcolor{green!20}{6-1-5} & 3-0-9 & 2-5-5 \\
\hline
\end{tabular}
\caption{Comparison with T2VTopA using Metric-1. A total of 126 prompts were evaluated, rather than 128, as T2VTopA rejected 2 prompts.} 
\label{ranking-hailuo}
\end{table}

\begin{table}[ht]\scriptsize
\centering
\begin{tabular}{c|c|c|c}
\hline
Step-Video-T2V vs. T2VTopB (Win-Tie-Loss) & Annotator-1  & Annotator-2 & Annotator-3 \\
\hline
\hline
Overall & 36-35-51 & \cellcolor{green!20}{67-10-45} & \cellcolor{green!20}{55-22-45} \\
\hline
\hline
Sports & \cellcolor{green!20}{8-2-2} & \cellcolor{green!20}{10-1-1} & \cellcolor{green!20}{8-2-2} \\
\hline
Food & 3-4-3 & \cellcolor{green!20}{7-1-2} & \cellcolor{green!20}{7-2-1} \\
\hline
Scenery & 2-6-4 & \cellcolor{yellow!20}{5-2-5} & \cellcolor{green!20}{5-4-3} \\
\hline
Animals & \cellcolor{yellow!20}{5-1-5} & 3-1-7 & 2-2-7 \\
\hline
Festivals & \cellcolor{green!20}{6-1-4} & \cellcolor{green!20}{6-0-5} & 2-4-5 \\
\hline
Combined Concepts & 1-4-7 & \cellcolor{green!20}{6-1-5} & 4-2-6 \\
\hline
Surreal & 2-0-6 & 3-0-5 & 2-1-5 \\
\hline
People & 1-3-7 & 4-1-6 & 3-1-7 \\
\hline
3D Animation & \cellcolor{green!20}{5-3-4} & \cellcolor{green!20}{11-0-1} & \cellcolor{green!20}{11-0-1} \\
\hline
Cinematography & 3-3-5 & 4-2-5 & 3-1-7 \\
\hline
Style & 0-8-4 & \cellcolor{green!20}{8-1-3} & \cellcolor{green!20}{8-3-1} \\
\hline
\end{tabular}
\caption{Comparison with T2VTopB using Metric-1. A total of 122 prompts were evaluated, rather than 128, as T2VTopB rejected 6 prompts.}
\label{ranking-kling}
\end{table}

\begin{table}[ht]\scriptsize
\centering
\resizebox{\textwidth}{!}{
\begin{tabular}{c|c|c|c|c|c}
\hline
 & Model & Instruction Following & Motion Smoothness & Physical Plausibility & Aesthetic Appeal \\
\hline
\hline
\multirow{3}*{Annotator-1} & Step-Video-T2V & 204 & \cellcolor{green!20}{210} & \cellcolor{green!20}{203} & 187 \\
~ & T2VTopA & \cellcolor{green!20}{211} & 200 & 198 & \cellcolor{green!20}{196} \\
~ & T2VTopB & 185 & 184 & 178 & 175 \\
\hline
\hline
\multirow{3}*{Annotator-2} & Step-Video-T2V & 211 & \cellcolor{green!20}{243} & \cellcolor{green!20}{256} & 217 \\
~ & T2VTopA & \cellcolor{green!20}{241} & \cellcolor{green!20}{243} & 242 & \cellcolor{green!20}{228} \\
~ & T2VTopB & 234 & 236 & 229 & 204 \\
\hline
\hline
\multirow{3}*{Annotator-3} & Step-Video-T2V & 170 & \cellcolor{green!20}{197} & \cellcolor{green!20}{172} & \cellcolor{green!20}{178} \\
~ & T2VTopA & \cellcolor{green!20}{177} & 177 & 153 & 171 \\
~ & T2VTopB & 164 & 163 & 139 & 148 \\
\hline
\hline
\multirow{3}*{Annotator-4}& Step-Video-T2V & 199 & \cellcolor{green!20}{232} & \cellcolor{green!20}{230} & \cellcolor{green!20}{225} \\
~ & T2VTopA & \cellcolor{green!20}{217} & 221 & 201 & 199 \\
~ & T2VTopB & 194 & 219 & 194 & 194 \\
\hline
\hline
\multirow{3}*{Annotator-5} & Step-Video-T2V & 218 & \cellcolor{green!20}{225} & \cellcolor{green!20}{213} & 211 \\
~ & T2VTopA & \cellcolor{green!20}{221} & 220 & \cellcolor{green!20}{213} & \cellcolor{green!20}{212} \\
~ & T2VTopB & 209 & 217 & 202 & 196 \\
\hline
\hline
\multirow{3}*{Annotator-6} & Step-Video-T2V & 187 & 213 & 251 & 211 \\
~ & T2VTopA & 193 & 201 & 259 & 197 \\
~ & T2VTopB & \cellcolor{green!20}{201} & \cellcolor{green!20}{224} & \cellcolor{green!20}{271} & \cellcolor{green!20}{227} \\
\hline
\end{tabular}
}
\caption{Comparison with T2VTopA and T2VTopB using Metric-2. We invited six human annotators to evaluate each video. For each evaluation dimension, we aggregated the scores given by each annotator across all prompts for that dimension. Prompts that were rejected by any model were excluded from the analysis for all models.}
\label{ranking-2}
\end{table}

From Table~\ref{ranking-hailuo}, Table~\ref{ranking-kling}, and Table~\ref{ranking-2} we got three observations.

First, the overall ranking of the three models in Table~\ref{ranking-hailuo} and Table~\ref{ranking-kling} is as follows: T2VTopA > Step-Video-T2V > T2VTopB. We analyzed categories such as Scenery, Animals, People, and Style, where Step-Video-T2V performs worse than the other two models, and found that the primary reason lies in their generally higher aesthetic appeal. We believe this advantage mainly stems from the higher resolutions of the generated videos (720P in T2VTopA, 1080P in T2VTopB, and 540P in Step-Video-T2V) and the high-quality aesthetic data used during their post-training stages. Table~\ref{ranking-2} also shows that 4 out of 6 annotators rate T2VTopA and T2VTopB as having higher aesthetic appeal.

Second, Step-Video-T2V consistently outperforms T2VTopA and T2VTopB in the Sports category in Table~\ref{ranking-hailuo} and Table~\ref{ranking-kling}, demonstrating its strong capability in modeling and generating videos with high-motion dynamics. Table~\ref{ranking-2} also highlights Step-Video-T2V's superiority in Motion Smoothness and Physical Plausibility.

Third, we observed that T2VTopA has better instruction-following capability, which contributes to its superior performance in categories such as Combined Concepts, Surreal, and Cinematography. We believe the key reasons for this are better video captioning model and the greater human effort involved in labeling the post-training data used by T2VTopA.

Note that Step-Video-T2V still lacks sufficient training in the final stage of pre-training with 540P videos, having only seen 25.3M samples (as shown in Table~\ref{trainingrecipe}). Additionally, compared to these two commercial engines, we are using significantly less high-quality data in the post-training phase, which will be continuously improved in the future. Finally, the video length is 204 frames, nearly twice the length of T2VTopA and T2VTopB, making our training more challenging. We assert that Step-Video-T2V has already achieved the strongest motion dynamics modeling and generation capabilities among all commercial engines. Given comparable training resources and high-quality data, we believe it can achieve state-of-the-art results in general domains as well.

We also compare Step-Video-T2V with the international commercial text-to-video engine, Runway Gen-3 Alpha, and present the results in Table~\ref{ranking-runway1} and Table~\ref{ranking-runway2}. Since Gen-3 Alpha has a stronger understanding of English, we translate the Chinese prompts into English before generating results. As shown in Table~\ref{ranking-runway1} and Table~\ref{ranking-runway2}, Step-Video-T2V outperforms Gen-3 Alpha overall, while Gen-3 Alpha excels in generating videos within the Cinematography domain.

\begin{table}[ht]\scriptsize
\centering
\begin{tabular}{c|c|c|c}
\hline
Step-Video-T2V vs. Gen-3 Alpha (Win-Tie-Loss) & Annotator-1  & Annotator-2 & Annotator-3 \\
\hline
\hline
Overall & \cellcolor{green!20}{68-3-38} & \cellcolor{green!20}{60-27-25} & \cellcolor{green!20}{54-36-22} \\
\hline
\hline
Sports & \cellcolor{green!20}{10-0-2} & \cellcolor{green!20}{11-1-0} & \cellcolor{green!20}{6-5-1} \\
\hline
Food & \cellcolor{green!20}{10-0-1} & \cellcolor{green!20}{7-2-2} & \cellcolor{green!20}{5-3-3} \\
\hline
Scenery & \cellcolor{green!20}{7-2-3} & \cellcolor{green!20}{7-2-3} & \cellcolor{green!20}{7-1-4} \\
\hline
Animals & \cellcolor{green!20}{7-1-4} & \cellcolor{green!20}{7-3-2} & 4-7-1 \\
\hline
Festivals & \cellcolor{green!20}{7-0-4} & \cellcolor{green!20}{6-5-0} & \cellcolor{green!20}{2-9-0} \\
\hline
Combined Concepts & \cellcolor{green!20}{6-1-5} & \cellcolor{yellow!20}{4-4-4} & \cellcolor{green!20}{9-0-3} \\
\hline
Surreal & 1-1-4 & 2-1-3 & \cellcolor{green!20}{6-0-0} \\
\hline
People & \cellcolor{green!20}{5-1-6} & \cellcolor{green!20}{5-3-4} & \cellcolor{green!20}{7-3-2} \\
\hline
3D Animation & \cellcolor{green!20}{1-0-0} & \cellcolor{green!20}{1-0-0} & \cellcolor{yellow!20}{0-1-0} \\
\hline
Cinematography & 4-0-7 & 2-3-6 & 3-3-5 \\
\hline
Style & \cellcolor{green!20}{10-0-2} & \cellcolor{green!20}{8-3-1} & \cellcolor{green!20}{5-4-3} \\
\hline
\end{tabular}
\caption{Comparison with Runway Gen-3 Alpha using Metric-1, excluding prompts that were rejected by Gen-3 Alpha.} 
\label{ranking-runway1}
\end{table}

\begin{table}[ht]\scriptsize
\centering
\resizebox{\textwidth}{!}{
\begin{tabular}{c|c|c|c|c|c}
\hline
 & Model & Instruction Following & Motion Smoothness & Physical Plausibility & Aesthetic Appeal \\
\hline
\hline
\multirow{2}*{Annotator-1} & Step-Video-T2V & \cellcolor{green!20}{214} & \cellcolor{green!20}{221} & \cellcolor{green!20}{214} & \cellcolor{green!20}{198} \\
~ & Gen-3 Alpha & 178 & 180 & 150 & 169 \\
\hline
\hline
\multirow{2}*{Annotator-2} & Step-Video-T2V & 183 & \cellcolor{green!20}{200} & \cellcolor{green!20}{210} & 173 \\
~ & Gen-3 Alpha & \cellcolor{green!20}{185} & 173 & 177 & \cellcolor{green!20}{176} \\
\hline
\hline
\multirow{2}*{Annotator-3} & Step-Video-T2V & 174 & \cellcolor{green!20}{202} & \cellcolor{green!20}{176} & 184 \\
~ & Gen-3 Alpha & \cellcolor{green!20}{179} & 180 & 158 & \cellcolor{green!20}{194} \\
\hline
\hline
\multirow{2}*{Annotator-4}& Step-Video-T2V & \cellcolor{green!20}{162} & \cellcolor{green!20}{186} & \cellcolor{green!20}{189} & \cellcolor{green!20}{183} \\
~ & Gen-3 Alpha & 147 & 165 & 133 & 160 \\
\hline
\hline
\multirow{2}*{Annotator-5} & Step-Video-T2V & \cellcolor{green!20}{228} & \cellcolor{green!20}{237} & \cellcolor{green!20}{225} & \cellcolor{green!20}{211} \\
~ & Gen-3 Alpha & 200 & 189 & 149 & 166 \\
\hline
\hline
\multirow{2}*{Annotator-6} & Step-Video-T2V & 160 & \cellcolor{green!20}{178} & \cellcolor{green!20}{207} & \cellcolor{green!20}{171} \\
~ & Gen-3 Alpha & \cellcolor{green!20}{178} & 161 & 182 & 153 \\
\hline
\end{tabular}
}
\caption{Comparison with Runway Gen-3 Alpha using Metric-2. We invited six human annotators to evaluate each video. For each evaluation dimension, we aggregated the scores given by each annotator across all prompts for that dimension. Prompts that were rejected by Gen-3 Alpha were excluded from the analysis for all models.}
\label{ranking-runway2}
\end{table}

\subsection{Evaluation on Movie Gen Video Bench}
Movie Gen Video Bench \cite{polyak2024moviegencastmedia} is another existing benchmark for the text-to-video generation task. It includes 1,003 prompts across multiple categories, covering human activities, animals, nature and scenery, physics, as well as unusual subjects and activities. Although Movie Gen Video has not been open-sourced, its generated results on the Movie Gen Video Bench are publicly available (\hyperlink{https://github.com/facebookresearch/MovieGenBench}{https://github.com/facebookresearch/MovieGenBench}). Therefore, we also compare Step-Video-T2V with Movie Gen Video and HunyuanVideo in Table~\ref{ranking-moviegenbench} using this benchmark.

\begin{table}[ht]\scriptsize
\centering
\resizebox{\textwidth}{!}{
\begin{tabular}{c|c|c|c}
\hline
Category & \makecell[c]{Step-Video-T2V vs. Movie Gen Video \\ (Win-Tie-Loss)} & \makecell[c]{Step-Video-T2V vs. HunyuanVideo \\ (Win-Tie-Loss)} & \# of Prompts \\
\hline
\hline
Overall & 485-315-489 & \cellcolor{green!20}{615-313-361} & 1,289 \\
\hline
\hline
human & 123-58-160 & \cellcolor{green!20}{181-64-96} & 341 \\
\hline
physics & 61-54-64 & \cellcolor{green!20}{87-47-45} & 179 \\
\hline
unusual activity \& subject & \cellcolor{green!20}{110-74-108} & \cellcolor{green!20}{136-75-81} & 292 \\
\hline
animal & 39-37-42 & \cellcolor{green!20}{47-30-41} & 118 \\
\hline
scene & \cellcolor{green!20}{84-53-6}3 & \cellcolor{green!20}{91-58-51} & 200 \\
\hline
sequential motion & \cellcolor{green!20}{9-2-2} & \cellcolor{green!20}{6-2-5} & 13 \\
\hline
camera motion & \cellcolor{green!20}{59-37-50} & \cellcolor{green!20}{67-37-42} & 146 \\
\hline
\end{tabular}
}
\caption{Comparison of Movie Gen Video and HunyuanVideo using the Movie Gen Video Bench. The total number of evaluations (1,289) is greater than 1,003 due to some prompts having multiple category tags. This evaluation involved six human annotators.}
\label{ranking-moviegenbench}
\end{table}

Compared to Movie Gen Video, Step-Video-T2V achieves a comparable performance.
We got several observations from this comparison.
First, the pre-training of Step-Video-T2V remains insufficient. While Movie Gen Video was trained on 73.8M videos during its high-resolution pre-training phase, Step-Video-T2V was trained on only 27.3M videos—about one-third of the number used by Movie Gen Video. Additionally, we observed that the training curves for all pre-training stages in Step-Video-T2V continue to show a downward trend. Due to resource limitations, we plan to conduct more extensive pre-training as part of our future work.
Second, the Movie Gen Video paper highlights the significant human effort involved in labeling the high-quality SFT dataset. However, due to limited human resources, we lack enough high-quality labeled data at this stage to effectively refine the visual style and quality of the generated results.
Third, Movie Gen Video can generate 720P videos, which are visually more appealing than the 540P resolution produced by Step-Video-T2V. Feedback from human annotators suggests that high resolution can often be a key factor in determining which model performs better.
Compared to HunyuanVideo, Step-Video-T2V achieves significant improvements across all categories, solidifying its position as the state-of-the-art open-source text-to-video model.

\subsection{Generating Text Content in Videos}
We also compare Step-Video-T2V with open-source and commercial engines on a list of prompts such as "\textit{a squirrel holding a sign that says 'hello'.}", where the model is required to generate videos that include text content as well.

Our observations show that Step-Video-T2V outperforms all other models in generating basic English text. We attribute this capability to the T2I pre-training stage, where a portion of the images contained text, and the captions explicitly described it. However, the accuracy of text generation remains far from ideal. Furthermore, due to the complexity of Chinese characters, Step-Video-T2V is currently able to generate only a limited number of them. Enhancing text generation capabilities for both English and Chinese will be a focus of our future work.

\begin{figure}[ht]
    \centering
    \includegraphics[width=1.3\textwidth, center, trim=0 0 0 0, clip]{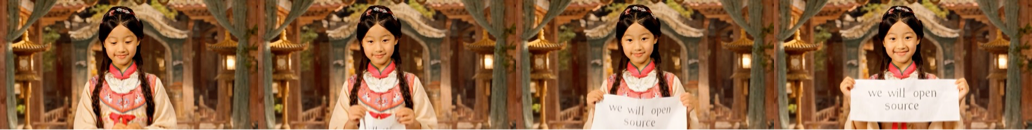}
    \caption{Four frames sampled from the video generated based on the prompt "\textit{In the video, a Chinese girl is dressed in an exquisite traditional outfit, smiling with a confident and graceful expression. She holds a piece of paper with the words "we will open source" clearly written on it. The background features an ancient and elegant setting, complementing the girl's demeanor. The entire scene is clear and has a realistic style.}".}
    \label{fig:text}
\end{figure}

\subsection{VAE Video Reconstruction}

\begin{table}[h]\scriptsize
    \centering
    \begin{tabular}{lcccc}
    \toprule
    \textbf{Model} & \textbf{Downsample Factor} & \textbf{SSIM$\uparrow$} & \textbf{PSNR$\uparrow$} & \textbf{rFVD$\downarrow$} \\
    \midrule
    OpenSora-1.2 (\cite{opensora}) & 4 × 8 × 8 & 0.9126 & 31.41 & 20.42 \\
    CogvideoX-1.5 (\cite{yang2024cogvideox}) & 4 × 8 × 8 & 0.9373 & 38.10 & 16.33 \\
    HunyuanVideo (\cite{kong2024hunyuanvideo}) & 4 × 8 × 8 & 0.9710 & \cellcolor{green!20}{39.56} & 4.17 \\
    Cosmos-VAE (\cite{agarwal2025cosmos}) & 4 × 8 × 8 & 0.9315 & 37.66 & 9.10 \\
    \hline
    Cosmos-VAE (\cite{agarwal2025cosmos}) & 8 × 16 × 16 & 0.8862 & 34.82 & 40.33 \\
    \hline
    Video-VAE (Ours) & 8 × 16 × 16 & \cellcolor{green!20}{0.9776} & 39.37 & \cellcolor{green!20}{3.61} \\
    \bottomrule
    \end{tabular}
    \caption{Comparison of reconstruction metrics.}
    \label{tab: vae}
\end{table}

We compare Video-VAE with several open-source baselines using 1,000 test videos from various domains, each with dimensions of 50(frames)×480(height)×768(width). As shown in Table~\ref{tab: vae}, despite having a compression ratio 8 times larger than most baselines, our reconstruction quality still maintains state-of-the-art performance. While Cosmos-VAE also offers a high-compression version with a factor of 8×16×16, its reconstruction quality falls significantly behind our method.

Figure~\ref{fig:vae} illustrates typical challenge cases in video reconstruction, including high-motion (first row), text (second row), texture (third row), high-motion combined with text (fourth row), and high-motion combined with texture (fifth row). Our models significantly outperform other baselines, even with higher compression ratios.
\begin{figure}[ht]
    \centering
    \includegraphics[width=1.3\textwidth, center, trim=0 0 0 0, clip]{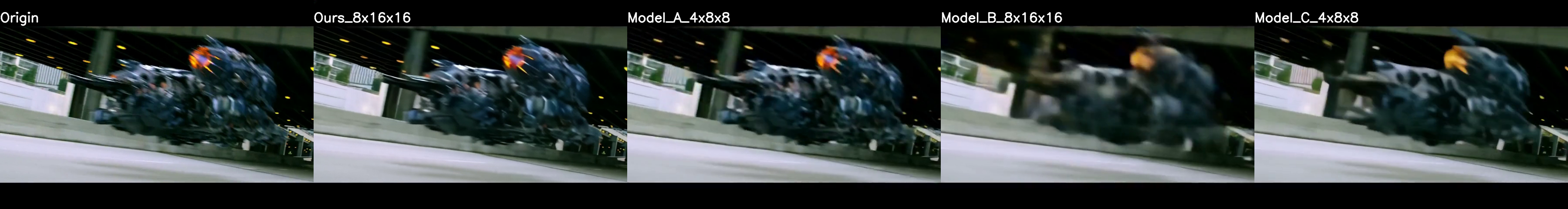}
    \includegraphics[width=1.3\textwidth, center, trim=0 0 0 0, clip]{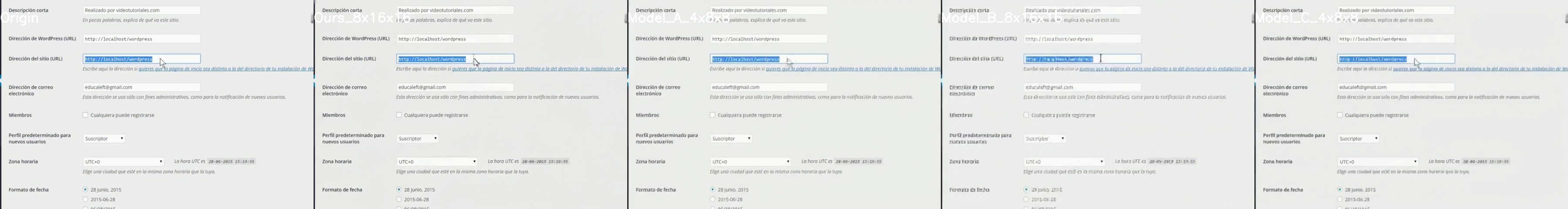}
    \includegraphics[width=1.3\textwidth, center, trim=0 0 0 0, clip]{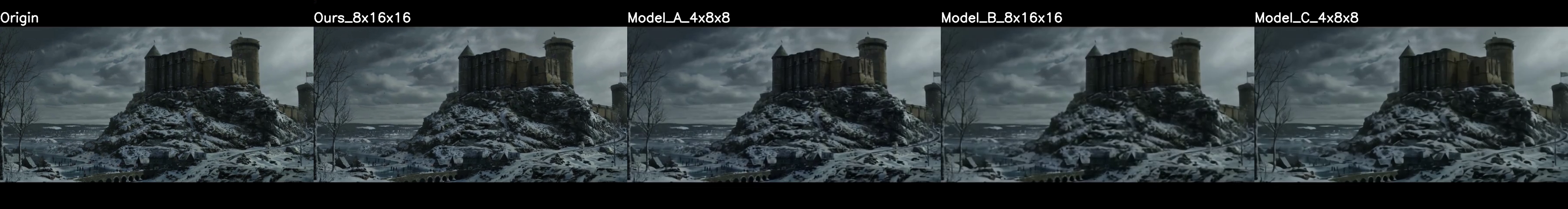}
    \includegraphics[width=1.3\textwidth, center, trim=0 0 0 0, clip]{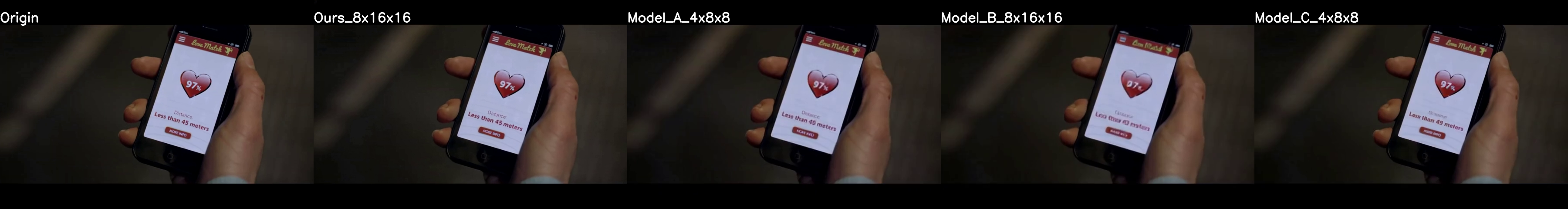}
    \includegraphics[width=1.3\textwidth, center, trim=0 0 0 0, clip]{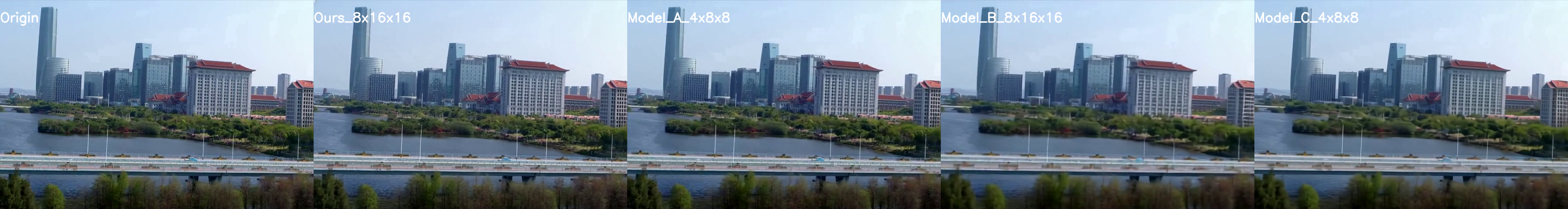}
    \caption{Video reconstruction results compared with public available models, in scenarios including high-motion (first row), text (second row), texture (third row), high-motion combined with text (fourth row), and high-motion combined with texture (fifth row).}
    \label{fig:vae}
\end{figure}
\subsection{DPO}
\begin{figure*}[ht]
    \centering
    \includegraphics[width=1.3\textwidth, center, trim=0 0 0 0, clip]{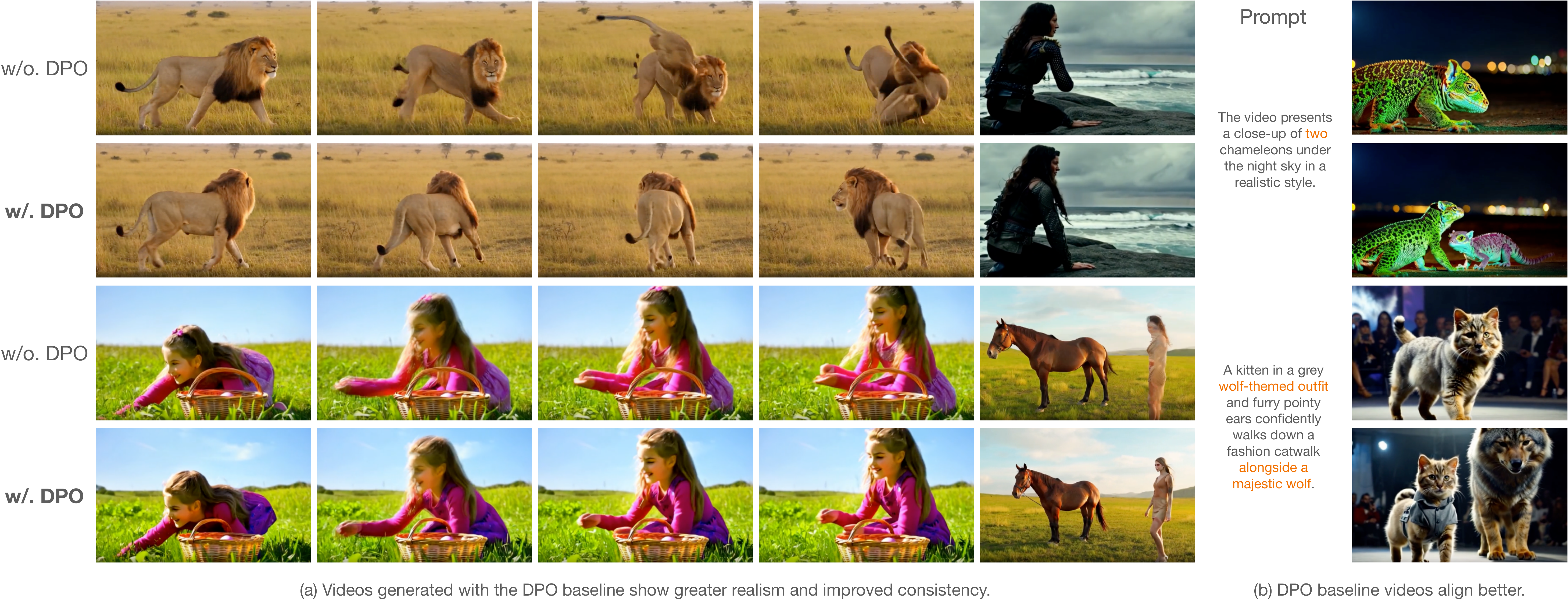}
    \caption{Visual comparison of video generation with and without the DPO baseline.}
    \label{fig:dpovisual}
\end{figure*}

To assess the effectiveness of the proposed Video-DPO algorithm, we conduct inference on 300 diverse prompts. The evaluation involves two models: the baseline model and the baseline model with the Video-DPO enhancement (baseline w/. DPO). Both models are sampled under identical initial noise conditions to control for extraneous variables and ensure a fair comparison. For each generated video, three independent annotators are tasked with evaluating their preference between the two models, with an option to select "no preference". The evaluation protocol is as follows: 
\begin{itemize}[left=0cm]
    \item If an annotator prefers the video generated by "baseline w/. DPO", the model receives 1 point.
    \item If an annotator prefers the "baseline" video, the baseline model receives 1 point.
    \item If an annotator indicates "no preference," both models receive 0.5 points.
\end{itemize}
Upon aggregating the scores, we find that the baseline model with DPO (baseline w/. DPO) achieves a preference score of 55\%, outperforming the baseline model (45\%). This result demonstrates the efficacy of Video-DPO in generating videos more aligned with user preferences. The visual comparison is shown in Figure~\ref{fig:dpovisual}, demonstrates that human feedback enhances the plausibility and consistency of generated videos. Additionally, we observe that the DPO baseline enhances the alignment with the given prompts, resulting in more accurate and relevant video generation. 

While Video-DPO demonstrates effectiveness, several issues remain. (1) The trajectory from initial noise to timestep-specific latents acts as implicit dynamic conditions beyond text prompts — yet this dimension remains underutilized due to computational limitations. (2)  A tradeoff exists between sparse and imprecise feedback, especially in video diffusion models. For instance, in videos with over 100 million pixels, only a few pixels may be problematic, yet feedback often comes as a single scalar or lacks precision. (3) Unlike LLMs, which use token-level softmax to create competition between tokens, diffusion models rely on regression, which may result in less efficient preference optimization. We hope these discussions provide insights and inspire further algorithmic advancements in incorporating human feedback.

\section{Discussion}
\subsection{Model Architecture}
Unlike DiT, which relies on a modulation mechanism to condition the network on the text prompt, MMDiT integrates the text prompt directly into the Transformer, separates the weights for text and video, and uses a shared attention mechanism to merge the latent representations of both modalities. We compared the training curves of DiT and MMDiT in the early stages and found both architectures exhibited similar performance. Given these comparable results, along with DiT's ability to disentangle text and video and its natural extension to pure video prediction models without text, we ultimately selected DiT as the model architecture for Step-Video-T2V. Due to computational cost constraints, we did not train MMDiT-based model for an extended period to assess its upper performance limit.

We also compared spatial-temporal attention and 3D full attention mechanisms within DiT. In the spatial-temporal attention mechanism, the model captures spatial information among tokens with the same temporal index within spatial Transformer blocks, and temporal information across time steps in temporal Transformer blocks. In contrast, 3D full attention mechanism combines both spatial and temporal information in a unified attention process, offering higher performance potential but at the cost of increased computational demands. We trained two DiT models—one using spatial-temporal attention and the other using 3D full attention—both in a 4B setting. Upon comparing their performances, we found that 3D full attention-based model outperforms spatial-temporal attention-based model, particularly in generating videos with high motion dynamics. Given its superior quality, we ultimately selected the 3D full attention setting.

In addition, 3D full attention is known for its high training and inference cost, so we are still actively investigating more efficient way to reduce the computation overhead, while preserving the same model quality~\cite{dsv}.

\subsection{Instruction Following}

Based on the evaluation results, we found that even a DiT-based model like Step-Video-T2V, with 30B parameters, struggles to generate videos involving complex action sequences. Additionally, generating videos that incorporate multiple concepts with low occurrence in the training data (e.g., an elephant and a penguin) remains challenging in Step-Video-T2V and other leading text-to-video generation models. Both of these challenges can be viewed as instruction-following problems.

We examine the instruction-following capability of Step-Video-T2V, focusing on how it interprets instructions involving various objects, actions, and other details. Our analysis reveals that the distribution of cross-attention scores is occasionally highly concentrated, with a strong focus on specific objects or actions. This pronounced attention can result in missing objects, wrong details, or incomplete action sequences in the generated videos.

By heuristically repeating the missing objects in the prompt, some of the problematic cases can be significantly improved. This demonstrates the importance of ensuring that all elements in the prompt receive appropriate attention. We leave the task of balancing this attention for future work, aiming to refine the model’s ability to better attend and follow all elements in the prompt.

\subsection{Laws of Physics Following}
We analyzed a number of videos generated by leading text-to-video models, including Sora, Hailuo, Kling, and Step-Video-T2V, and found that all of these models struggle to accurately simulate the real world and generate videos that adhere to the laws of physics—such as \textit{a ball bouncing on the floor} or \textit{a drop of water falling into a cup}. Some text-to-video engines can produce good results for certain prompts, but these successes are often due to the model over-fitting to specific annotations, and cannot generalize well. 

This finding highlights a key limitation of diffusion-based models in text-to-video generation. To address this challenge, we plan to develop more advanced model paradigms in future work, such as combining autoregressive and diffusion models within a unified framework (\cite{chen2024diffusionforcingnexttokenprediction}, \cite{hacohen2024ltxvideorealtimevideolatent}, \cite{zhou2025tamingteacherforcingmasked}), to better adhere to the laws of physics and more accurately simulate realistic interactions.

\subsection{High-quality Labeled Data for Post-training}
By applying a small amount of high-quality human-labeled data in SFT, Step-Video-T2V achieves significant improvements in the overall video quality, demonstrating that the quality and diversity of the data outweigh its sheer scale. We also observed that certain characteristics of these curated high-quality datasets, such as video style and the degree of motion dynamics, generalize well across a broader range of prompts. This further underscores the importance of high-quality, small-scale, and diverse datasets for post-training.

Curating such datasets is both expensive and time-consuming, involving tasks such as selecting high-quality videos from a large pool, labeling them with accurate captions, and ensuring the dataset covers a diverse range of objects, actions, styles, and domains. We plan to build a comprehensive video knowledge base with structured labels as part of our future work.

\subsection{RL-based Optimization Mechanism for Post-training}
We employed a simple yet effective DPO-based model for video generation and also explored training a reward model to automate the entire post-training process. However, the proposed method still requires human labeling efforts in the early stages and is time-consuming when extending it to general domains. On the other hand, RL-based approaches have achieved great success in LLMs, such as OpenAI-O1 and DeepSeek-R1 \cite{deepseekai2025deepseekr1incentivizingreasoningcapability}. However, unlike RL-focused natural language tasks, such as solving math problems or generating code, which have well-defined problems with clear answers, it remains challenging to define similar tasks in the video generation domain. We consider this a key challenge for future research exploration.

\section{Conclusion and Future Work}

This technical report introduces and open-sources Step-Video-T2V, a state-of-the-art pre-trained video generation model from text, featuring 30B parameters, a deep compression Video-VAE, a DPO approach for video generation, and the ability to generate videos up to 204 frames in length. We provide a comprehensive overview of our pre-training and post-training strategies and introduce Step-Video-T2V-Eval as a new benchmark for evaluating text-to-video generation models.

We highlight challenges faced by current text-to-video models. First, high-quality labeled data remains a significant hurdle. Existing video captioning models often struggle with hallucination issues, and human annotations are expensive and difficult to scale. Second, instruction-following requires more attention, as it encompasses a wide range of scenarios, from generating videos based on detailed descriptions to handling complex action sequences and combinations of multiple concepts. Third, current models still face difficulties in generating videos that obey the laws of physics, an issue stemming from the inherent limitations of diffusion models. Lastly, RL-based optimization mechanisms are areas worth exploring for post-training improvements in video generation models.

Looking ahead, we plan to launch a series of open-source projects focused on the development of video foundation models, starting with Step-Video-T2V. We hope these efforts will drive innovation in video foundation models and empower video content creators.

{
\small
\bibliographystyle{unsrtnat}
\bibliography{main}
}

\newpage
\section*{Contributors and Acknowledgments}\label{team}

We designate core contributors as those who have been involved in the development of Step-Video-T2V throughout its entire process, while contributors are those who worked on the early versions or contributed part-time. All contributors are listed in \textbf{alphabetical order by first name}.

\begin{itemize}[left=0cm] 
\item \textbf{Core Contributors:}
    \begin{itemize}
        \item \textbf{Model \& Training:} Guoqing Ma, Haoyang Huang, Kun Yan, Liangyu Chen, Nan Duan, Shengming Yin.
        \item \textbf{Infrastructure:} Changyi Wan, Ranchen Ming, Xiaoniu Song, Xing Chen, Yu Zhou, Yuchu Luo.
        \item \textbf{Data \& Evaluation:} Deshan Sun, Deyu Zhou, Jian Zhou, Jianjian Sun, Kaijun Tan, Kang An, Liang Zhao, Mei Chen, Wei Ji, Qiling Wu, Wen Sun, Xin Han, Yanan Wei, Zheng Ge.
    \end{itemize}

\item \textbf{Contributors:} Aojie Li, Bin Wang, Bizhu Huang, Bo Wang, Brian Li, Changxing Miao, Chen Xu, Chenfei Wu, Chenguang Yu, Dapeng Shi, Dingyuan Hu, Enle Liu, Gang Yu, Ge Yang, Guanzhe Huang, Gulin Yan, Haiyang Feng, Hao Nie, Haonan Jia, Hanpeng Hu, Hanqi Chen, Haolong Yan, Heng Wang, Hongcheng Guo, Huilin Xiong, Huixin Xiong, Jiahao Gong, Jianchang Wu, Jiaoren Wu, Jie Wu, Jie Yang, Jiashuai Liu, Jiashuo Li, Jingyang Zhang, Junjing Guo, Junzhe Lin, Kaixiang Li, Lei Liu, Lei Xia, Liang Zhao, Liguo Tan, Liwen Huang, Liying Shi, Ming Li, Mingliang Li, Muhua Cheng, Na Wang, Qiaohui Chen, Qinglin He, Qiuyan Liang, Quan Sun, Ran Sun, Rui Wang, Shaoliang Pang, Shiliang Yang, Shuli Gao, Sitong Liu, Siqi Liu, Song Yuan, Tiancheng Cao, Tianyu Wang, Weipeng Ming, Wenqing He, Wuxun Xie, Xu Zhao, Xuelin Zhang, Xianfang Zeng, Xiaojia Liu, Xuan Yang, Yanbo Yu, Yang Li, Yaoyu Wang, Yaqi Dai, Yineng Deng, Yingming Wang, Yilei Wang, Yuanwei Lu, Yu Chen, Yu Luo, Yuanhao Ding, Yuhe Yin, Yuheng Feng, Yuxiang Yang, Zecheng Tang, Zekai Zhang, Zidong Yang.

\item \textbf{Project Sponsors:} Binxing Jiao, Daxin Jiang, Heung-Yeung Shum, Jiansheng Chen, Jing Li, Shuchang Zhou, Xiangyu Zhang, Xinhao Zhang, Yibo Zhu.

\item \textbf{Corresponding Authors:} Daxin Jiang (djiang@stepfun.com), Nan Duan (nduan@stepfun.com).

\end{itemize}

\end{document}